\pdfoutput=1

\documentclass{article}
\usepackage[letterpaper,top=2cm,bottom=2cm,left=3cm,right=3cm,marginparwidth=1.75cm]{geometry}

\usepackage{algorithm}
\usepackage{algorithmic}
\usepackage{amsmath}
\usepackage{svg}
\usepackage{arydshln}
\usepackage{amssymb}
\usepackage{booktabs} 
\usepackage{colortbl}
\usepackage{graphicx} 
\usepackage{hhline}
\usepackage[colorlinks=true,linkcolor=blue,citecolor=blue]{hyperref}
\usepackage{latexsym}
\usepackage{makecell}
\usepackage{multirow}
\usepackage{natbib}
\usepackage{tabu}
\usepackage{times}
\usepackage{wrapfig}
\usepackage{xcolor}
\usepackage{mathtools}
\DeclarePairedDelimiter\ceil{\lceil}{\rceil}

\newcommand{\bw}{\mathbf{w}}
\newcommand{\btheta}{\boldsymbol{\theta}}
\newcommand{\gr}{\rowcolor[gray]{.95}}

\usepackage[T1]{fontenc}

\usepackage[utf8]{inputenc}

\usepackage{microtype}

\usepackage{inconsolata}

\definecolor{lightblue}{HTML}{E8F0FE}
\NewDocumentCommand{\chihan}{ mO{} }{\textcolor{blue}{\textsuperscript{\textit{Chi}}{#1}}}

\NewDocumentCommand{\heng}{ mO{} }{\textcolor{red}{\textsuperscript{\textit{Heng}}{#1}}}

\NewDocumentCommand{\shizhe}
{ mO{} }{\textcolor{blue}{\textsuperscript{\textit{shizhe}}\textsf{\textbf{\small[#1]}}}}

\DeclareSymbolFont{extraup}{U}{zavm}{m}{n}
\DeclareMathSymbol{\vardiamond}{\mathalpha}{extraup}{87}

\title{LISA: Layerwise Importance Sampling for Memory-Efficient Large Language Model Fine-Tuning}

\author{\bf Rui Pan$^{\spadesuit*}$, ~ \bf Xiang Liu$^{\clubsuit*}$ ~ \bf Shizhe Diao$^{\vardiamond}$, ~ \bf Renjie Pi$^{\heartsuit}$, ~ \bf Jipeng Zhang$^{\heartsuit}$, \\~ \bf Chi Han$^{\spadesuit}$, ~ \bf Tong Zhang$^{\spadesuit}$\\
  $^{\spadesuit}$University of Illinois Urbana-Champaign\\
  $^{\clubsuit}$The Hong Kong University of Science and Technology(Guangzhou)\\
  $^{\vardiamond}$NVIDIA $^{\heartsuit}$The Hong Kong University of Science and Technology\\
  \texttt{\{ruip4, chihan3, tozhang\}@illinois.edu}\\
  \texttt{xliu886@connect.hkust-gz.edu.cn} \\
  \texttt{shizhe.diao@gmail.com} \quad \texttt{\{rpi, jzhanggr\}@ust.hk}
  \\
  \\
}

\begin{document}
\maketitle
\def\thefootnote{*}\footnotetext{Equal Contribution.}
\begin{abstract}
The machine learning community has witnessed impressive advancements since large language models (LLMs) first appeared. Yet, their massive memory consumption has become a significant roadblock to large-scale training. For instance, a 7B model typically requires at least 60 GB of GPU memory with full parameter training, which presents challenges for researchers without access to high-resource environments. Parameter Efficient Fine-Tuning techniques such as Low-Rank Adaptation (LoRA) have been proposed to alleviate this problem. However, in most large-scale fine-tuning settings, their performance does not reach the level of full parameter training because they confine the parameter search to a low-rank subspace. Attempting to complement this deficiency, we investigate the layerwise properties of LoRA on fine-tuning tasks and observe an unexpected but consistent skewness of weight norms across different layers. Utilizing this key observation, a surprisingly simple training strategy is discovered, which outperforms both LoRA and full parameter training in a wide range of settings with memory costs as low as LoRA. We name it \textbf{L}ayerwise \textbf{I}mportance \textbf{S}ampled \textbf{A}damW (\textbf{LISA}), a promising alternative for LoRA, which applies the idea of importance sampling to different layers in LLMs and randomly freeze most middle layers during optimization. Experimental results show that with similar or less GPU memory consumption, LISA surpasses LoRA or even full parameter tuning in downstream fine-tuning tasks, where LISA consistently outperforms LoRA by over $10\%$-$35\%$ in terms of MT-Bench score while achieving on-par or better performance in MMLU, AGIEval and WinoGrande. On large models, specifically LLaMA-2-70B, LISA surpasses LoRA on MT-Bench, GSM8K, and PubMedQA, demonstrating its effectiveness across different domains.
\end{abstract}

\section{Introduction}
Large language models (LLMs) like ChatGPT excel in tasks such as writing documents, generating complex code, answering questions, and conducting human-like conversations~\citep{ouyang2022instructgpt}. With LLMs being increasingly applied in diverse task domains, domain-specific fine-tuning has emerged as a critical strategy to enhance their downstream capabilities~\citep{raffel2020t5, Chowdhery2022PaLMSL, rozière2023code, openai2023gpt4}. 
Nevertheless, these methods are typically time-intensive and consume substantial computational resources, posing significant challenges to the development of large-scale models. For example, continual pre-training typically requires several weeks even with multiple 80 GB GPUs.
To reduce costs, Parameter-Efficient Fine-Tuning (PEFT) techniques have been proposed to minimize the number of trainable parameters. These techniques include adapter weights~\citep{houlsby2019peft}, prompt weights~\citep{li2021prefix}, and LoRA~\citep{hu2022lora}. Among these, LoRA stands out as one of the most widely adopted due to its unique ability to merge the adaptor back into the base model parameters, significantly enhancing efficiency. However, LoRA's superior performance in fine-tuning tasks has yet to reach a point that universally surpasses full parameter fine-tuning in all settings~\citep{ding2022delta,dettmers2023qlora}. In particular, it has been observed that LoRA tends to falter on large-scale datasets during continual pre-training~\citep{lialin2023relora}, which raises doubts about the effectiveness of LoRA under those circumstances. 
\begin{wrapfigure}{r}{0.44\textwidth}
    \vspace{-10pt}
    \begin{center}
        \includegraphics[width=1\linewidth]{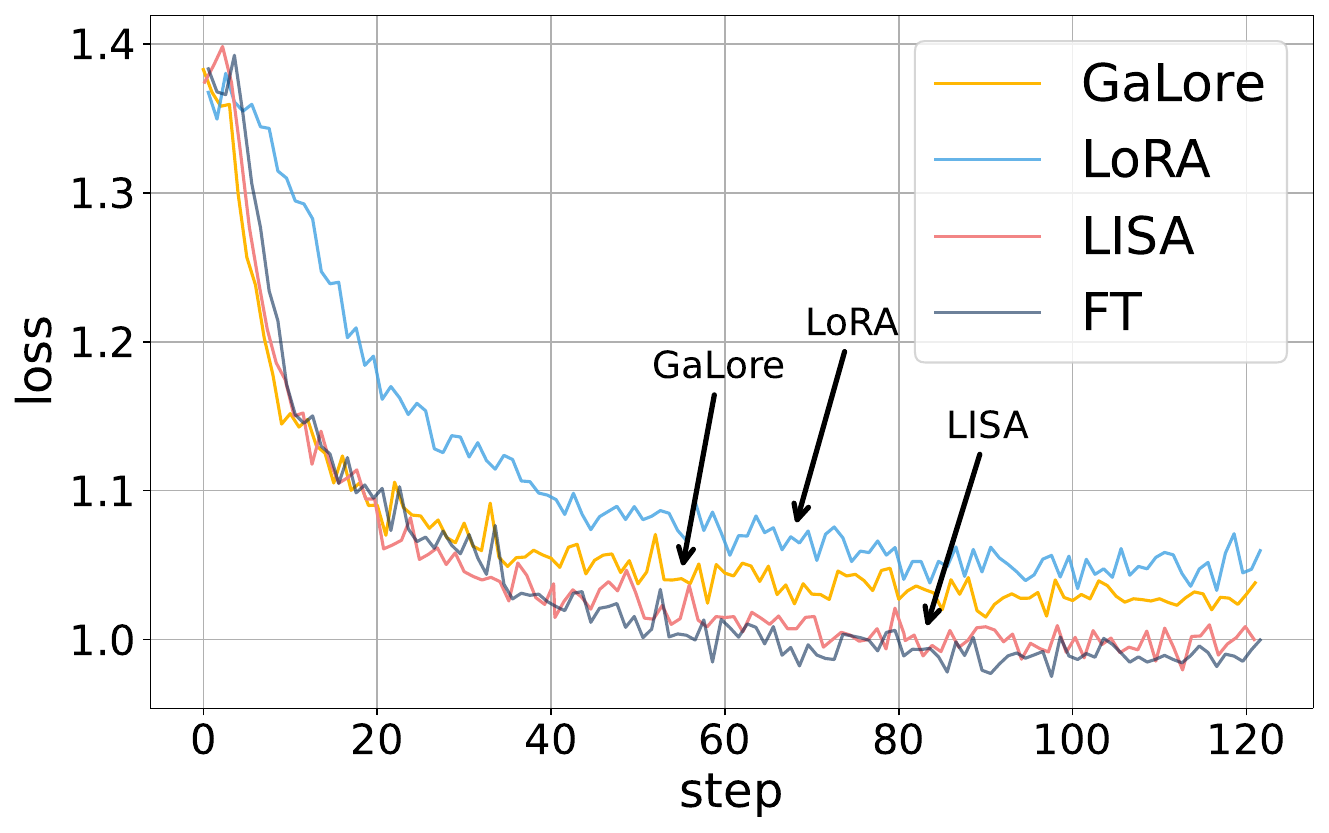}
        \caption{Training loss of LLaMA-2-7B model on Alpaca GPT-4 dataset with Full Parameter Training (FT), LoRA, GaLore, and LISA.}
        \label{fig:gpt2_con_pre_c4}
    \end{center}
    \vspace{-10pt}  
\end{wrapfigure}
We attribute this to LoRA's much fewer trainable parameters compared to the base model, which limits the representation power of LoRA training.

To overcome this shortcoming, we delve into LoRA's training statistics in each layer, aspiring to bridge the difference between LoRA and full-parameter fine-tuning. Surprisingly, we discover that LoRA's layerwise weight norms have an uncommonly skewed distribution, where the bottom layer and/or the top layer occupy the majority of weights during the update. In contrast, the other self-attention layers only account for a small amount, which means different layers have different importance when updating. This key observation inspires us to ``sample'' different layers by their importance, which matches the idea of importance sampling~\citep{kloek1978bayesian,zhao2015stochastic}.

As a natural consequence, this strategy brings forth our \textbf{L}ayerwise \textbf{I}mportance \textbf{S}ampled \textbf{A}dam (\textbf{LISA}) algorithm, where by selectively updating only essential LLM layers and leaving others untouched, LISA enables training large-scale language models ($\ge 65$B parameters) with less or similar memory consumption as LoRA. Furthermore, fine-tuned on downstream tasks, LISA outperformed both LoRA and conventional full-parameter fine-tuning approaches by a large margin, indicating the large potential of LISA as a promising alternative to LoRA. 

We summarize our key contributions as follows,

\begin{itemize}
  \item We discover the phenomenon of skewed weight-norm distribution across layers in LoRA, which implies the varied importance of different layers in large-scale LLM training.
  \item We propose the Layerwise Importance Sampled AdamW (LISA), a simple optimization method capable of scaling up to over $70$B LLMs with less or similar memory cost as LoRA.
  \item We demonstrate LISA's effectiveness in fine-tuning tasks for modern LLMs, where it outperforms LoRA by $10\%$-$35\%$ in  MT-Bench and achieves better performance in multiple benchmarks. In addition, LISA exhibits much better convergence behaviors than LoRA. LISA even outperforms full parameters training under certain settings. Similar performance gain is observed across different sized models ($7$B-$70$B) and tasks, including instruction following, medical QA, and math problems.
\end{itemize}
\section{Related Work}

\subsection{Large Language Models}
In the realm of natural language processing (NLP), the Transformer architecture has been a revolutionary technique, initially known for its effectiveness in machine translation tasks~\citep{Vaswani_Shazeer_Parmar_Uszkoreit_Jones_Gomez_Kaiser_Polosukhin_2017}. With the inception of models like BERT~\citep{devlin-etal-2019-bert} and GPT-2~\citep{radford2019gpt2}, the approach shifted towards pre-training on extensive corpora, which led to significant performance enhancements in downstream fine-tuning tasks~\citep{raffel2020t5,brown_language_2020,zhang2022opt,workshop2022bloom,falcon180b,touvron2023llama,touvron2023llama2,vicuna2023,biderman2023pythia,jiang2024mixtral}. However, the growing number of parameters in these models results in a huge GPU memory consumption, rendering the fine-tuning of large scale models ($\ge 65$B) infeasible under low resource scenarios. This has prompted a shift towards more efficient training of LLMs.

\subsection{Parameter-Effieient Fine-Tuning}
Parameter-efficient fine-tuning (PEFT) methods adapt pre-trained models by fine-tuning only a subset of parameters. In general, PEFT methods can be grouped into three classes: 1) Prompt Learning methods~\citep{li2021prefix,hambardzumyan2021warp, zhong2021factual, han2021ptr, qin2021learning, liu2021gpt, diao2022black}, 2) Adapter methods~\citep{houlsby2019peft,hu2022lora,diao2021taming,diao2023mixture,meng2024pissa,song2024increasing,ji2024advlora}, and 3) Selective methods~\citep{liu2021autofreeze,liu2021autofreeze,li2023smartfrz, luo2024badam}. 
Prompt learning methods emphasize optimizing the input token or input embedding with frozen model parameters, which generally has the least training cost among all three types. 
Adapter methods normally introduce an auxiliary module with much fewer parameters than the original model, and updates are only applied to the adapter module during training.
Compared with them, selective methods are more closely related to LISA, which focuses on optimizing a fraction of the model's parameters without appending extra modules. 
Recent advances in this domain have introduced several notable techniques through layer freezing. AutoFreeze~\citep{liu2021autofreeze} offers an adaptive mechanism to identify layers for freezing automatically and accelerates the training process. FreezeOut~\citep{brocker2017freezeout} progressively freezes intermediate layers, significantly reducing training time without notably affecting accuracy. The SmartFRZ~\citep{li2023smartfrz} framework utilizes an attention-based predictor for layer selection, substantially cutting computation and training time while maintaining accuracy. However, none of these layer-freezing strategies has been widely adopted in the context of Large Language Models due to their inherent complexity or non-compatibility with modern memory reduction techniques~\citep{rajbhandari2020zero, rasley2020deepspeed} for LLMs.

\subsection{Low-Rank Adaptation (LoRA)}
In contrast, the Low-Rank Adaptation (LoRA) technique is much more prevalent in common LLM training~\citep{hu2022lora}. LoRA reduces the number of trainable parameters by employing low-rank matrices, thereby lessening the computational burden and memory cost. One key strength of LoRA is its compatibility with models featuring linear layers, where the decomposed low-rank matrices can be merged back into the original model. This allows for efficient deployment without changing the model architecture. As a result, LoRA can be seamlessly combined with other techniques, such as quantization~\citep{dettmers2023qlora} or Mixture of Experts~\citep{gou2023mixture}. Despite these advantages, LoRA's performance is not universally comparable with full parameter fine-tuning. There have been tasks in~\citep{ding2022delta} that LoRA performs much worse than full parameter training on. This phenomenon is especially evident in large-scale pre-training settings~\citep{lialin2023relora}, where to the best of our knowledge, only full parameter training was adopted for successful open-source LLMs~\citep{falcon180b,touvron2023llama,touvron2023llama2,jiang2023mistral,zhang2024tinyllama,jiang2024mixtral}.

\subsection{Large-scale Optimization Algorithms}
In addition to approaches that change model architectures, there have also been efforts to improve the efficiency of optimization algorithms for LLMs.
One such approach is layerwise optimization, a concept with roots extending back several decades. Notably, \citep{Hinton_Osindero_Teh_2006} introduced an effective layer-by-layer pre-training method for Deep Belief Networks (DBN), demonstrating the benefits of sequential layer optimization. This idea was expanded by researchers like \citep{Bengio_Lamblin_Popovici_Larochelle_2007}, who illustrated the advantages of a greedy, unsupervised approach to pre-training each layer of deep networks. In the context of large batch training, \citep{you2017lars,you2019lamb} developed LARS and LAMB to improve generalization and mitigate the performance declines associated with large batch sizes. Despite these innovations, Adam~\citep{kingma2014adam,reddi2019amsgrad} and AdamW~\citep{loshchilov2017adamw} continue to be the predominant optimization methods used in most LLM settings.

Recently, other attempts have also been made to reduce the training cost of LLMs. For example, MeZO~\citep{malladi2023finetuning} adopted zeroth order optimization, bringing significant memory savings during training. However, it also incurred a considerable performance drop in multiple benchmarks, particularly in complex fine-tuning scenarios. Regarding acceleration, Sophia~\citep{liu2023sophia} incorporates clipped second-order information into the optimization, obtaining non-trivial speedup on LLM training. The significant downsides are its intrinsic complexity of Hessian estimation and unverified empirical performance in large-size models (e.g., $\ge 65$B). In parallel to our work,~\citep{zhao2024galore} proposed GaLore, a memory-efficient training strategy that reduces memory cost by projecting gradients into a low-rank compact space. Yet the performance has still not surpassed full-parameter training in fine-tuning settings. To sum up, LoRA-variant methods~\citep{hu2022lora,dettmers2023qlora, zhao2024galore} with AdamW~\citep{loshchilov2017adamw} is still the dominant paradigm for large-size LLM fine-tuning, the performance of which still demands further improvements.


\section{Method}

\subsection{Motivation}
To understand how LoRA achieves effective training with only a few parameters, we conducted empirical studies on multiple models, especially observing the weight norms across various layers. We fine-tune it on the Alpaca-GPT4 dataset \citep{peng2023instruction}. During the training, we meticulously recorded the mean weight norms of each layer $\ell$ at every step $t$ after updates, i.e.

\begin{align*}
  \bw^{(\ell)} \triangleq \texttt{mean-weight-norm}(\ell) = \frac{1}{T} \sum_{t=1}^{T} \|\btheta^{(\ell)}_t\|_2
\end{align*}

\begin{figure}[!ht]
\begin{center}
\includegraphics[width=0.44\linewidth]{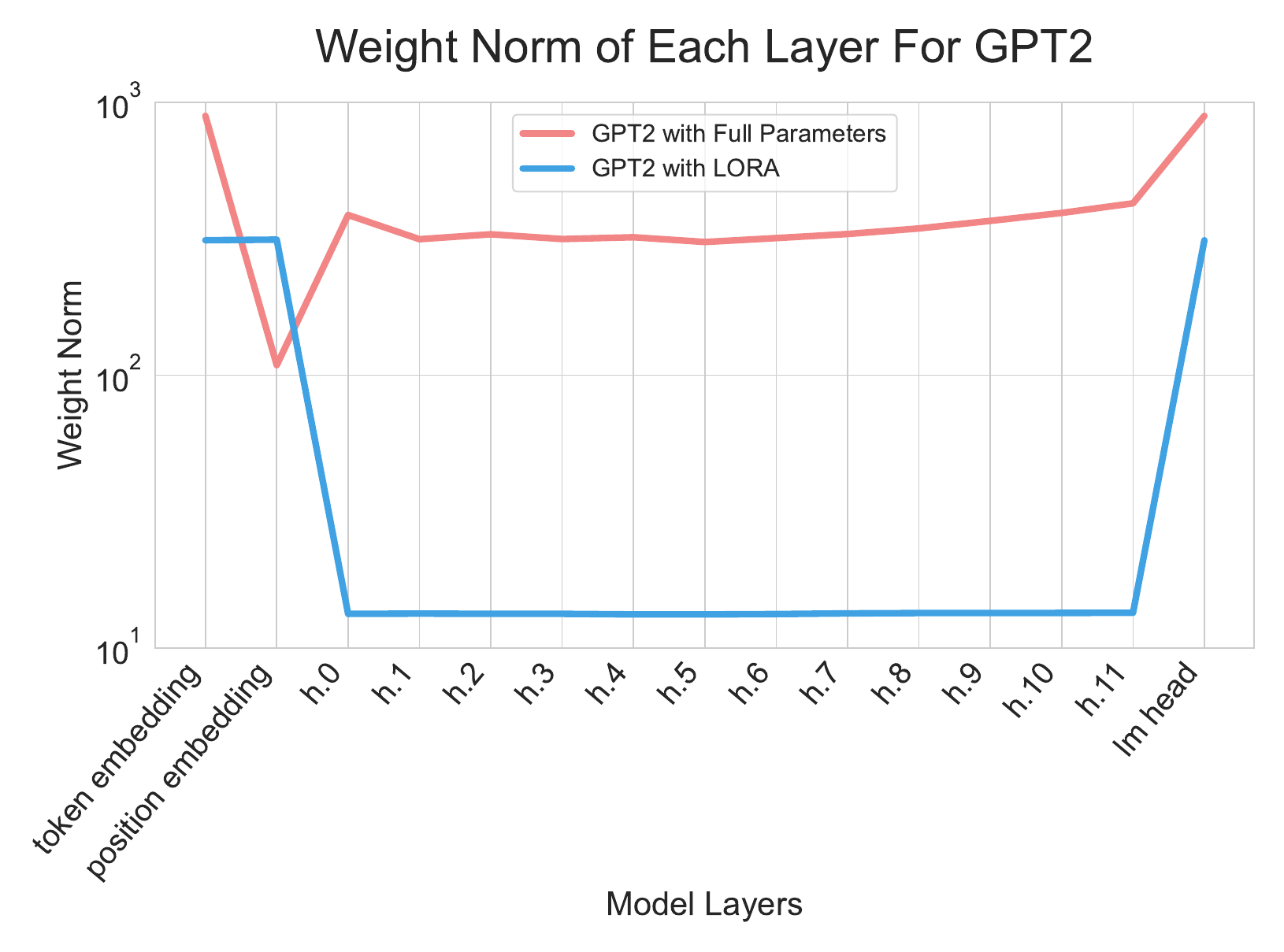}
\includegraphics[width=0.45\linewidth]{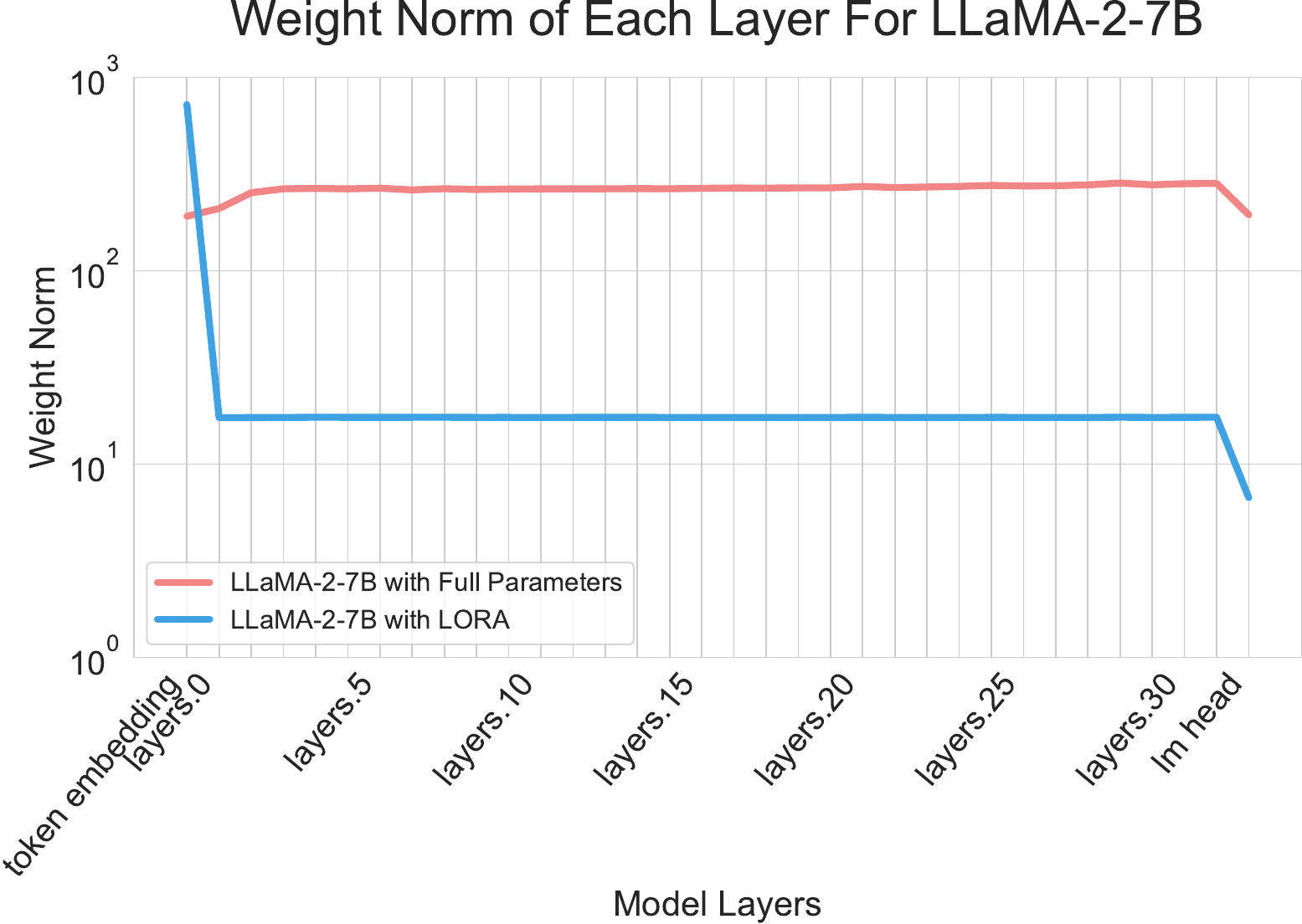}
\caption{Layer-wise weight norms during training of GPT2 and LLaMA-2-7B Model with LoRA and Full Parameters training.} 
\label{fig:GPT2_LoRA_fp_weight_norm}
\end{center}
\end{figure}

Figure~\ref{fig:GPT2_LoRA_fp_weight_norm} presents these findings, with the x-axis representing the layer id, from embedding weights to the final layer, and the y-axis quantifying the weight norm. The visualization reveals one key trend:
\begin{itemize}
\item The embedding layer or the language modeling (LM) head layer exhibits significantly larger weight norms than intermediary layers in LoRA, often by a factor of hundreds. This phenomenon, however, was not salient under full-parameter training settings. 
\end{itemize}

This observation indicates that the update emphasis of LoRA and full parameter training differ significantly, which can be attributed to the difference in their learned knowledge. For example, in embedding layers, tokens with similar meanings, i.e., synonyms, can be projected into the same embedding space and converted to similar embeddings. LoRA may capture this similarity in language and ``group'' them in the low-dimension space, allowing frequent features of language meanings to be promptly identified and optimized. The price is LoRA's limited representation power restricted by its intrinsic low-rank space, as we can see from the comparison with LISA in image generation tasks (Appendix~\ref{appendix:stable_diffusion}), where LoRA memorizes and learns details much slower than LISA. Other possible explanations can also justify this phenomenon. Despite various interpretations of this observation, one fact remains clear: \textit{LoRA values layerwise importance differently from full parameter tuning}.

\subsection{Layerwise Importance Sampled AdamW (LISA)}

To exploit the discovery above, we aspire to simulate LoRA's updating pattern via sampling different layers to freeze. This way, we can avoid LoRA's inherent deficiency of limited low-rank representation ability and emulate its fast learning process. Intuitively, given the same global learning rates across layers, layers with small weight norms in LoRA should also have small sampling probabilities to unfreeze in full-parameter settings so the expected learning rates across iterations can stay the same. This is exactly the idea of importance sampling~\citep{kloek1978bayesian,zhao2015stochastic}, where instead of applying layerwise different learning rates $\{\eta_t\}$ in full-parameter settings to emulate LoRA's updates $\{\tilde{\eta}_t\}$, we apply sampling and instead  get the same expected parameter update 
\begin{align*}
  \eta^{(\ell)}_t = \tilde{\eta}^{(\ell)}_t \cdot \frac{\tilde{\bw}^{(\ell)}}{\bw^{(\ell)}}
  \quad\Rightarrow\quad
  \eta^{(\ell)}_t = \eta^{(\ell)}, p^{(\ell)} = \frac{\tilde{\bw}^{(\ell)}}{\bw^{(\ell)}}
\end{align*}
This gives rise to our Layerwise Importance Sampling AdamW method, as illustrated in Algorithm~\ref{alg:lisa}. In practice, since all layers except the bottom and top layer have small weight norms in LoRA, we adopt $\{p_\ell\}_{\ell=1}^{N_L} = \{1.0, \gamma/N_L, \gamma/N_L, \dots, \gamma/N_L, 1.0\}$ in practice, where $\gamma$ controls the expected number of unfreeze layers during optimization. Intuitively, $\gamma$ serves as a compensation factor to bridge the difference between LoRA and full parameter tuning, letting LISA emulate a similar layerwise update pattern as LoRA. To further control the memory consumption in practical settings, we instead randomly sample $\gamma$ layers every time to upper-bound the maximum number of unfrozen layers during training.

\begin{algorithm}
\caption{\textbf{L}ayerwise \textbf{I}mportance \textbf{S}ampling \textbf{A}damW (\textbf{LISA})}
\label{alg:lisa}
\begin{algorithmic}[1]
\REQUIRE number of layers $N_L$, number of iterations $T$, sampling period $K$, number of sampled layers $\gamma$, initial learning rate $\eta_0$
\FOR{$i \gets 0$ to $T/K - 1$}
    \STATE Freeze all layers except the embedding and language modeling head layer
    \STATE Randomly sample $\gamma$ intermediate layers to unfreeze
    \STATE Run AdamW for $K$ iterations with $\{\eta_t\}_{t=ik}^{ik+k-1}$
\ENDFOR
\end{algorithmic}
\end{algorithm}
\section{Experimental Results}
\label{sec:exp}
\begin{table*}[!h]
\caption{The chart illustrates peak GPU memory consumption for various model architectures and configurations, highlighting differences across models. The LISA configuration is specifically labeled in the table: ``E'' denotes the embedding layer, ``H'' represents the language modeling head layer, and ``2L'' indicates two additional intermediate layers. *: Model parallelism is applied for the 70B model.}
\begin{center}
\begin{sc}
\begin{small}
    \resizebox{0.85\linewidth}{!}{
    \begin{tabular}{l|r|rrr|rrr}
        \toprule[1pt]
                   & Vanilla & \multicolumn{3}{c|}{LoRA rank} & \multicolumn{3}{c}{LISA activate layers} \\ \midrule
        
        Model      & -       & 128       & 256       & 512    & E+H         & E+H+2L       & E+H+4L      \\ 
        \midrule
        GPT2-Small & 3.8G     & 3.3G       & 3.5G       & 3.7G    & 3.3G         & 3.3G          & 3.4G         \\
        TinyLlama  & 13G     & 7.9G      & 8.6G      & 10G    & 7.4G        & 8.0G         & 8.3G        \\
        Mistral-7B & 59G     & 23G       & 26G       & 28G    & 21G         & 23G          & 24G         \\
        LLaMA-2-7B  & 59G     & 23G       & 26G       & 28G    & 21G         & 23G          & 24G         \\ 
        LLaMA-2-70B* & OOM &   79G & OOM & OOM & 71G & 75G & 79G \\
        \bottomrule
    \end{tabular}
}
\end{small}
\end{sc}
\end{center}
\label{tab:peak_gpu_memory}
\end{table*}

\subsection{Memory Efficiency}
\label{sec:memory_efficiency}

We conducted peak GPU memory experiments to demonstrate LISA's memory efficiency and showcase its comparable or lower memory cost than LoRA.

\paragraph{Settings}

To reasonably estimate the memory cost, we randomly sample prompts from the Alpaca dataset~\citep{alpaca} and limit the maximum output token length to 1024. 
We focus on two key hyperparameters: LoRA's rank and LISA's number of activation layers. 
For other hyperparameters, a mini-batch size of 1 was consistently used across five LLMs from 120M to 70B parameters, deliberately excluding other GPU memory-saving techniques such as gradient checkpointing~\citep{chen2016training}, offloading~\citep{ren2021zerooffload}, and flash attention~\citep{dao2022flashattention, dao2023flashattention2}.

\begin{wrapfigure}{r}{0.45\textwidth}
\vspace{-10pt}
\begin{center}
\includegraphics[width=1\linewidth]{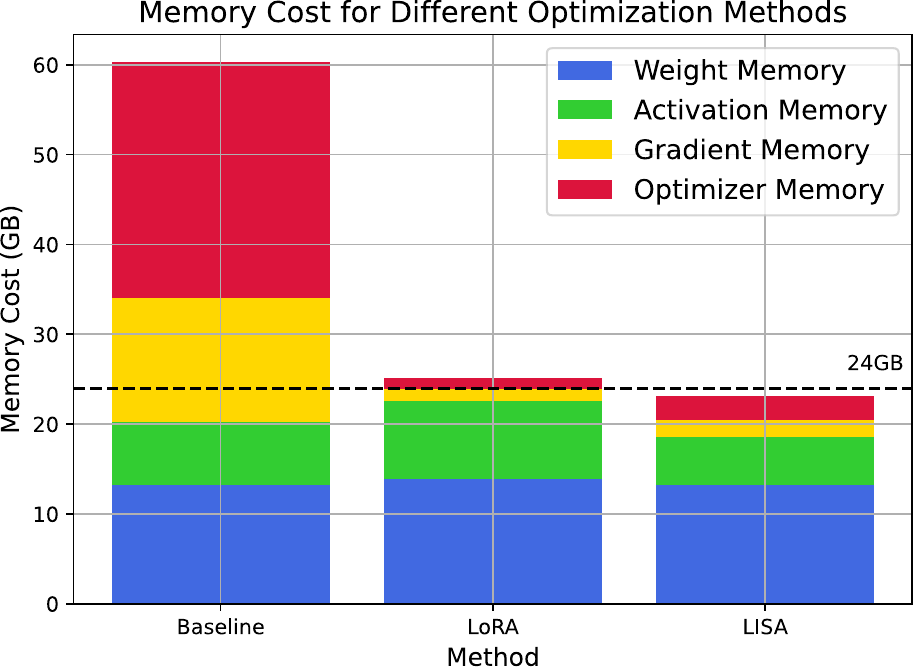}
\caption{GPU memory consumption of LLaMA-2-7B with different methods and batch size 1.}
\label{fig:memory_cost_of_llama-2-7b}
\end{center}
\end{wrapfigure}

All memory-efficiency experiments are conducted on 4$\times$ NVIDIA Ampere Architecture GPUs with 80G memory. 

\paragraph{Results}
Upon examining Table~\ref{tab:peak_gpu_memory}, it is evident that the LISA configuration, particularly when enhanced with both the embedding layer (E) and two additional layers (E+H+2L), demonstrates a considerable reduction in GPU memory usage when fine-tuning the LLaMA-2-70B model, as compared to the LoRA method. Specifically, the LISA E+H+2L configuration shows a decrease to 75G of peak GPU memory from the 79G required by the LoRA Rank 128 configuration. This efficiency gain is not an isolated incident; a systematic memory usage decrease is observed across various model architectures, suggesting that LISA's method of activating layers is inherently more memory-efficient.

In Figure~\ref{fig:memory_cost_of_llama-2-7b}, it is worth noticing that the memory reduction in LISA allows LLaMA-2-7B to be trained on a single RTX4090 (24GB) GPU, which makes high-quality fine-tuning affordable even on a laptop computer. In particular, LISA requires much less activation memory consumption than LoRA since it does not introduce additional parameters brought by the adaptor. LISA's activation memory is even slightly less than full parameter training since pytorch~\citep{paszke2019pytorch} with deepspeed~\citep{rasley2020deepspeed} allows deletion of redundant activations before backpropagation.

\begin{wrapfigure}{r}{0.45\textwidth}
\vspace{-10pt}  
\begin{center}
\includegraphics[width=1\linewidth]{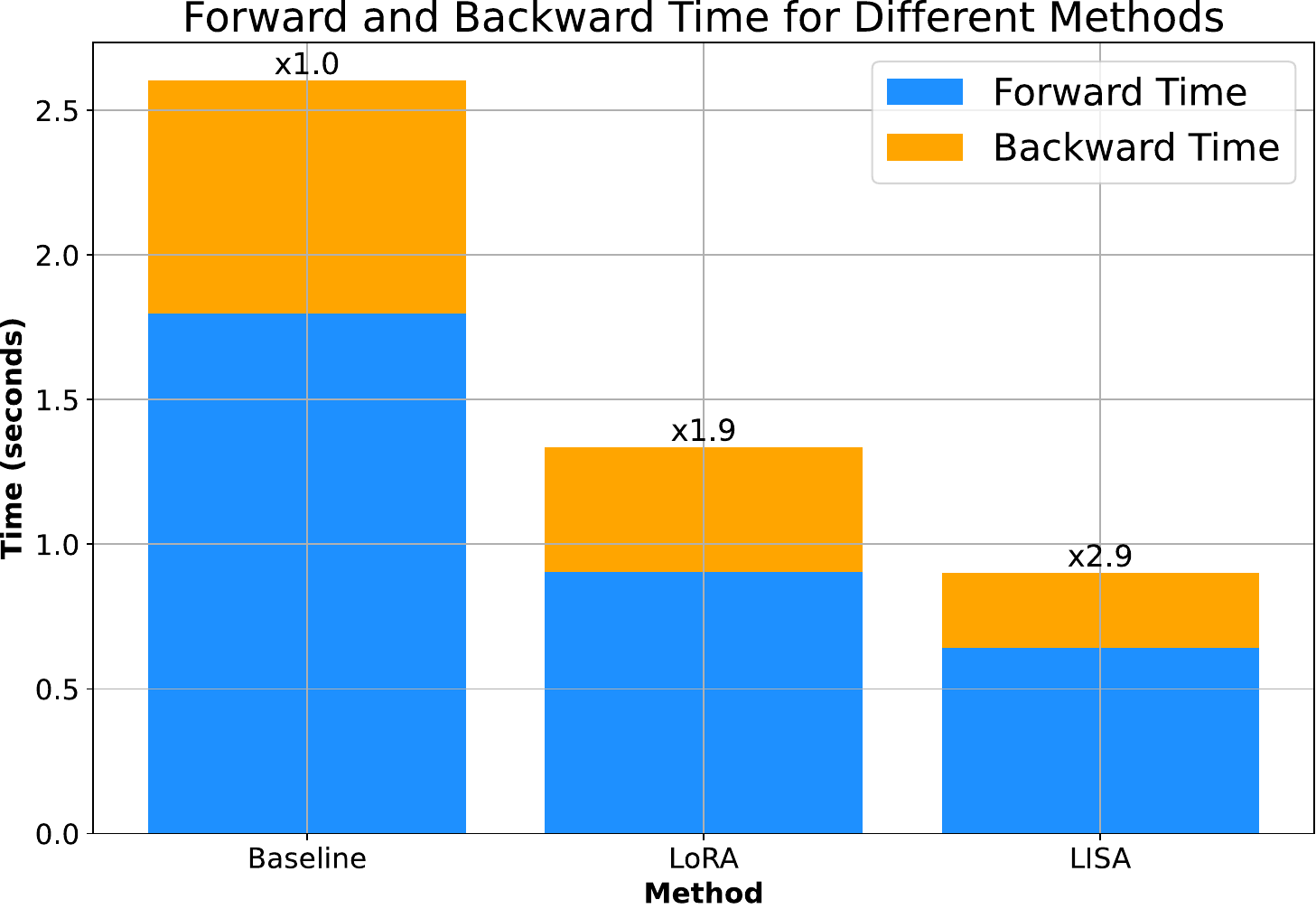}
\caption{Single-iteration time cost of LLaMA-2-7B with different methods and batch size 1.}
\vspace{-25pt}
\label{fig:time_cost_of_llama-2-7b}
\end{center}
\end{wrapfigure}

On top of that, a reduction in memory footprint from LISA also leads to an acceleration in speed. As shown in Figure~\ref{fig:time_cost_of_llama-2-7b}, LISA provides almost $2.9\times$ speedup when compared with full-parameter training, and $\sim1.5\times$ speedup against LoRA, partially due to the removal of adaptor structures. It is worth noticing that the reduction of memory footprint in both LoRA and LISA leads to a significant acceleration of forward propagation, emphasizing the importance of memory-efficient training.

\subsection{Moderate Scale Fine-Tuning}
\label{sec:exp_moderate_ft}

LISA can achieve this significant memory saving while still obtaining competitive performance under the fine-tuning setting.

\begin{table*}[!ht]
\vspace{10pt}
\caption{Results of different methods on MMLU, AGIEval, and WinoGrande, measured by accuracy.}
\begin{center}
\begin{sc}
\centering
\small
\begin{tabular}{ll|rrr}
\toprule
Model & Method & MMLU (5-shot) & AGIEval (3-shot) & WinoGrande (5-shot) \\ \midrule
\multirow{5}[1]{*}{TinyLlama} 
&Vanilla
& 25.50
& 19.55
& 59.91 \\

& LoRA
& 25.81
& 19.82
& 61.33 \\

& GaLore
&  25.21
&  21.19
&  61.09\\

& \textbf{LISA} 
& \textbf{26.02}
& \textbf{21.71}
& 61.48 \\

\gr \cellcolor{white}  &  FT
& 25.62
& 21.28
& \textbf{62.12} \\ \midrule

\multirow{5}[1]{*}{Mistral-7B} 
& Vanilla
& 60.12 
& 26.79 
& 79.24 \\

& LoRA
& 61.78 
& 27.56
& 78.85 \\

& GaLore
& 57.87 
& 26.23 
& 75.85 \\

& \textbf{LISA}
& \textbf{62.09}
& \textbf{29.76}
& \textbf{78.93} \\ 

\gr \cellcolor{white} &  FT
& 61.70 
& 28.07 
& 78.85 \\ \midrule

\multirow{5}[1]{*}{LLaMA-2-7B} 
& Vanilla
& 45.87
& 25.69
& 74.11 \\

& LoRA
& 45.50
& 24.73 
& 74.74 \\

& GaLore
& 45.56
& 24.39
& 73.32 \\

& \textbf{LISA}
& \textbf{46.21} 
& 26.06
& \textbf{75.30} \\

\gr  \cellcolor{white} &  FT
& 45.66 
& \textbf{27.02}
& 75.06 \\ \bottomrule
\end{tabular}
\end{sc}
\end{center}
\label{tab:all_main_full}
\end{table*}

\paragraph{Settings}
To demonstrate the superiority of LISA over LoRA, we evaluate them on the instruction-following fine-tuning task with the Alpaca GPT-4 dataset~\citep{alpaca}, which consists of 52k conversation pairs generated by GPT-4~\citep{openai2023gpt4}. The effectiveness of fine-tuning was evaluated on multiple benchmarks: MT-Bench~\citep{zheng2023judging} features 80 high-quality, multi-turn questions designed to assess LLMs on multiple aspects; MMLU~\citep{hendrycks2020measuring} includes a total of 57 tasks with 14,079 questions covering a broad spectrum of world knowledge; AGIEval~\citep{zhong2023agieval} serves as a human-centric benchmark for general abilities, comprising 9,316 instances; WinoGrande~\citep{sakaguchi2021winogrande} is a large-scale dataset for commonsense reasoning, consisting of 44,000 instances designed to challenge models' understanding of the context and commonsense knowledge.

In our experiments, we assessed three baseline models: TinyLlama~\citep{zhang2024tinyllama}, Mistral-7B~\citep{jiang2023mistral}, and LLaMA-2-7B~\citep{touvron2023llama2}. These models, varying in size ranging from 1B to 7B parameters, provide a diverse representation of decoder-only models. For hyper-parameters, we adopt a rank of 128 for LoRA and E+H+2L for LISA in this section, with full details available in Appendix~\ref{appendix:train_setup_and_hyperparame}.

\begin{wraptable}{r}{0.44\textwidth}
\footnotesize
\caption{Different methods on MT-Bench.}
\begin{center}
\begin{sc}
\centering
\resizebox{\linewidth}{!}{
\begin{tabular}{ll|c}
\toprule
Model & Method &  MT-Bench $\uparrow$\\ \midrule
\multirow{5}[1]{*}{TinyLlama}
& Vanilla
& 1.25 \\

& LoRA
& 1.90 \\

& GaLore
& \textbf{2.61} \\

& \textbf{LISA} 
& 2.57 \\

\gr \cellcolor{white} & FT
& 2.21 \\ \midrule

\multirow{5}[1]{*}{Mistral-7B} 
& Vanilla
& 4.32 \\

&  LoRA   
& 4.41 \\

& GaLore
& 4.36 \\

& \textbf{LISA}
& \textbf{4.85} \\

\gr \cellcolor{white} & FT
& 4.64 \\ \midrule

\multirow{5}[1]{*}{LLaMA-2-7B} 
& Vanilla
& 3.29 \\

& LoRA
& 4.45 \\

& GaLore
& 4.63 \\

& \textbf{LISA}
& \textbf{4.94} \\

\gr \cellcolor{white} & FT
& 4.75 \\ \bottomrule
\end{tabular}
}
\end{sc}
\end{center}
\label{tab:mt_bench_scores}
\vspace{-10pt}
\end{wraptable}

\paragraph{Results}
Table\ref{tab:all_main_full} and \ref{tab:mt_bench_scores} present a detailed comparison on moderate-scale LLMs. The baselines include Full-parameter Training (FT), Low-Rank Adaptation (LoRA)~\citep{hu2022lora} and  Gradient Low-Rank Projection (GaLore)~\citep{zhao2024galore}. The results demonstrate that LISA consistently outperforms other fine-tuning methods in most evaluation tracks, indicating its robustness and effectiveness across diverse tasks and model architectures. LISA is particularly effective in instruction following tasks, where a large gap is observed when compared with other baseline methods. LISA even outperforms Full-parameter Training, suggesting that an implicit regularization effect is present when the number of unfrozen layers is restricted, which is similar to dropout~\citep{JMLR:v15:srivastava14a:dropout}. According to more results in stable diffusion and detailed MT-Bench scores, we found that LISA outperforms LoRA mostly in memorization tasks, such as depicting high-resolution image details in image generation, or Writing or Humanities tasks in instruction following. This implies that LISA's performance improvement may majorly come from the ability to memorize long-tailed patterns, while LoRA is better at multi-hop reasoning with limited knowledge. For more details, please refer to Appendix \ref{appendix:stable_diffusion} and \ref{app:ft}.

\subsection{Moderate Scale Continual Pre-training}
\label{sec:exp_con_pretrain}

Continual pre-training is crucial for enabling models to adapt to new data and domains. To evaluate LISA's efficacy in the continual pre-training scenario, we experiment on the mathematics domain in comparison with Full-parameter Training.

\paragraph{Settings}
We adopt the mathematics corpus OpenWebMath~\citep{paster2023openwebmath} for constructing the continual pre-training dataset. Specifically, we extracted a high-quality subset from it which contains 1.5 billion tokens. Full details are explained in Appendix~\ref{app:con_data}. After continual pre-trainig, we then apply the same fine-tuning procedure on the GSM8K~\citep{cobbe2021training} training set, which comprises 7473 instances.

\begin{wraptable}{r}{0.5\textwidth}
\footnotesize
\vspace{-11pt}
\caption{Comparison of Moderate Scale Model Continual Pre-training on OpenWebMath Dataset.}
\begin{sc}
\begin{center}
\centering
\resizebox{\linewidth}{!}{
        \begin{tabular}{ll|cr }
        \toprule
        Model& Method   & GSM8K $\uparrow$  & Mem. $\downarrow$ \\ \midrule
        \multirow{3}[1]{*}{TinyLlama}
        & Vanilla 
        & 2.26 
        & - \\
    
        
        & \textbf{LISA}
        & \textbf{3.56} 
        & \textbf{8G} \\ 
        
        \gr \cellcolor{white} &  FT
        & 3.26 
        & 13G \\ \midrule
        
        \multirow{3}[1]{*}{LLaMA-2-7B}
        & Vanilla 
        & 14.40 
        & - \\

        
        & \textbf{LISA}
        & \textbf{22.21} 
        & \textbf{26G} \\ 

        \gr \cellcolor{white}  & FT
        & 22.21  
        & 59G \\ \midrule
        \end{tabular}
    }
\end{center}
\vspace{-10pt}
\end{sc}
\label{tab:7B_con_pretrain_OWM}
\end{wraptable}

\paragraph{Results}
Table \ref{tab:7B_con_pretrain_OWM} shows that LISA is capable of achieving on-par or even better performance than full-parameter training with much less memory consumption. Specifically, LISA requires only half of the memory cost compared to full-parameter training. This indicates a better balance between computational efficiency and model performance is achieved by LISA. According to our experience, reducing the number of unfrozen layers to half the original size leads to no worse or even better performance during continual pretraining, while requiring much less memory consumption.

\subsection{Large Scale Fine-Tuning}
\label{sec:exp_large_ft}

\begin{wraptable}{r}{0.52\textwidth}
\vspace{-23pt}
\caption{Different methods on MT-Bench, GSM8K, and PubMedQA score for LLaMA-2-70B.}
\begin{sc}
\begin{center}
\centering
    \resizebox{\linewidth}{!}{
        \begin{tabular}{l|ccc}
        \toprule
          Method       & MT-Bench$\uparrow$ & GSM8K$\uparrow$  & PubMedQA$\uparrow$ \\ \midrule
        Vanilla & 5.19 &  54.8 & 83.0\\
        LoRA    &  6.10 & 59.4 & 90.8\\
        \textbf{LISA} & \textbf{6.72}  &    61.1 & \textbf{91.6} \\ 
        \gr FT      &  6.25 &  \textbf{67.1} & 90.8\\ \bottomrule
        \end{tabular}
    }
\end{center}
\vspace{-20pt}
\end{sc}
\label{tab:70B_domain}
\end{wraptable}

To further demonstrate LISA's scalability on large-sized LLMs, we conduct additional fine-tuning experiments on LLaMA-2-70B~\citep{touvron2023llama2}.

\paragraph{Settings} On top of the aforementioned instruction-following tasks in Section~\ref{sec:exp_moderate_ft}, we use extra domain-specific fine-tuning tasks on mathematics and medical QA benchmarks. The GSM8K dataset~\citep{cobbe2021training}, comprising 7473 training instances and 1319 test instances, is used for the mathematics domain. 
For the medical domain, we select the PubMedQA dataset~\citep{jin2019pubmedqa}, which includes 211.3K artificially generated QA training instances and 1K test instances.

\begin{wraptable}{r}{0.5\textwidth}
\vspace{-5pt}
\caption{Different LISA hyperparameters combinations. All settings adopt learning rate $\eta_0=10^{-5}$. Here $\gamma$ stands for sampling layers, $K$ stands for sampling period. }
\vskip 0.15in
\begin{sc}
\centering
    \resizebox{\linewidth}{!}{
    \begin{tabular}{crc|c}
        \toprule
                  Models  & $\gamma$ & $K$
                   & \makecell{MT-Bench\\Score}
                  \\ \midrule

\multirow{8}[1]{*}{TinyLlama}
& \multirow{4}[1]{*}{2}
& $\ceil{T/125}$
& 2.44
\\
  
& 
& $\ceil{T/25}$
& \textbf{2.73} 
\\

& 
& $\ceil{T/5}$
& 2.64 
\\

&  
& $T$
& 2.26
\\
\cmidrule{2-4}

& \multirow{4}[1]{*}{8}
& $\ceil{T/125}$
& 2.59
\\

& 
& $\ceil{T/25}$ 
& \textbf{2.81}
\\ 

& 
& $\ceil{T/5}$
& 2.74
\\

& 
& $T$
& 2.53
\\
\midrule

\multirow{8}[1]{*}{LLaMA-2-7B}
& \multirow{4}[1]{*}{2}
& $\ceil{T/125} $
& 4.86
\\

& 
& $\ceil{T/25} $
& \textbf{4.91}
\\
 
& 
& $\ceil{T/5}$
& 4.88
\\
  
& 
& $T$
& 4.64
\\ 
\cmidrule{2-4}

& \multirow{4}[1]{*}{8} 
& $\ceil{T/125}$
& 4.94
\\
 
&  
& $\ceil{T/25}$
& \textbf{5.11}
\\

&  
& $\ceil{T/5}$
& 5.01
\\

&  
& $T$
& 4.73
\\
\bottomrule
        \end{tabular}
    }

\label{tab:all_ablation}
\vspace{-30pt}
\end{sc}
\end{wraptable}

Evaluation on the PubMedQA dataset~\citep{jin2019pubmedqa} is conducted in a 5-shot prompt setting, while the GSM8K dataset~\citep{cobbe2021training} assessment was conducted using Chain-of-Thought (CoT) prompting, following recent studies~\citep{wei2022chain, shum2023automatic, diao2023active}. 
Regarding hyperparameters, as detailed in the section \ref{sec:memory_efficiency}, we utilize the rank 256 for LoRA and the configuration  E+H+4L for LISA. Further information is available in Appendix~\ref{appendix:train_setup_and_hyperparame}.

\paragraph{Results}
As shown in Table~\ref{tab:70B_domain}, LISA consistently produces better or on-par performance when compared with LoRA. Furthermore, LISA again surpasses full-parameter training in instruction-tuning tasks, providing strong evidence to support LISA's scalability under large-scale training scenarios. More results are available in Appendix~\ref{app:ft}.

\subsection{Ablation Studies}

\paragraph{Hyperparameters of LISA}
The two key hyperparameters of LISA are the number of sampling layers $\gamma$ and sampling period $K$. To obtain intuitive and empirical guidance of those hyperparameter choices, we conduct ablation studies using TinyLlama~\citep{zhang2024tinyllama} and LLaMA-2-7B~\citep{touvron2023llama2} models with the Alpaca-GPT4 dataset. The configurations for $\gamma$, such as {E+H+2L, E+H+8L}, were denoted as $\gamma=2$ and $\gamma=8$. As for the sampling period  $K=T/n$, $T=122$ representing the maximum training step within our experimental framework. The findings, presented in Table~\ref{tab:all_ablation}, reveal that both $\gamma$ and $K$ markedly affect the LISA algorithm's performance. Specifically, a higher $\gamma$ value increases the quantity of trainable parameters, albeit with higher memory costs. On the other hand, an optimal $K$ value facilitates more frequent layer switching, thereby improving performance to a certain threshold, beyond which the performance may deteriorate. Generally, the rule of thumb is: \textit{More sampling layers and higher sampling period lead to better performance}. For a detailed examination of loss curves and MT-Bench results, refer to Appendix~\ref{app:ablation}.

\begin{wraptable}{r}{0.5\textwidth}
\vspace{-0.2 in}
\caption{The MT-Bench scores derived from varying random seeds for layer selection.}
\begin{sc}
\centering
    \resizebox{\linewidth}{!}{
        \begin{tabular}{c|ccc}
        \toprule
        Model      & Seed 1 &  Seed 2 &  Seed 3 \\ \midrule
        TinyLlama & 2.57         & 2.55  & 2.60 \\
        Mistral-7B & 4.85       & 4.82   & 4.82  \\
        LLaMA-2-7B & 4.94         &  4.92 & 4.89\\
        \bottomrule
        \end{tabular}
    }
    \vskip 0.1in
\label{tab:random_ablation}
\end{sc}
\end{wraptable}

\paragraph{Sensitiveness of LISA}
As LISA is algorithmically dependent on the sampling sequence of layers, it is intriguing to see how stable LISA's performance is under the effect of randomness. For this purpose, we further investigate LISA's performance variance over three distinct runs, each with a different random seed for layer selection. Here, we adopt TinyLlama, LLaMA-2-7B, and Mistral-7B models with the Alpaca-GPT4 dataset while keeping all other hyperparameters consistent with those used in the instruction following experiments in section~\ref{sec:exp_moderate_ft}. As shown in Table~\ref{tab:random_ablation}, LISA is quite resilient to different random seeds, where the performance gap across three runs is within $0.13$, a small value compared to the performance gains over baseline methods. For more ablation experiment on LISA hyperparameters, please refer to Appendix \ref{app:ablation}.

\section{Discussion}

\paragraph{Theoretical Properties of LISA}
\label{theorm}
\newtheorem{theorem}{Theorem}

Compared with LoRA, which introduces additional parameters and leads to changes in loss objectives, layerwise importance sampling methods enjoy nice convergence guarantees in the original loss. For layerwise importance sampled SGD, similar to gradient sparsification~\citep{wangni2018gradient}, the convergence can still be guaranteed for unbiased estimation of gradients with increased variance. The convergence behavior can be further improved by reducing the variance with appropriately defined importance sampling strategy~\citep{zhao2015stochastic}. For layerwise importance sampled Adam, theoretical results in~\citep{zhou2020randomized} prove its convergence in convex objectives. If we denote $f$ as the loss function and assume that the stochastic gradients are bounded, then based on~\citep{loshchilov2017adamw}, we know that AdamW optimizing $f$ aligns with Adam optimizing $f$ with a scaled regularizer, which can be written as 
\begin{align*}
    f^{\rm reg}(\bw) \triangleq f(\bw) + \frac{1}{2} \bw^\top \mathbf{S} \bw ,
\end{align*}
where $\mathbf{S}$ is a finite positive semidefinite diagonal matrix.
Following existing convergence results of RBC-Adam~(Corollary 1 in~\citep{zhou2020randomized}), we have the convergence guarantee of LISA in Theorem~\ref{thm:convergence_lisa}.

\begin{theorem}\label{thm:convergence_lisa}
    Let the loss function $f$ be convex and smooth. If the algorithm runs in a bounded convex set and the stochastic gradients are bounded, the sequence $\{\bw_t\}_{t=1}^T$ generated by LISA admits the following convergence rate:
    \begin{align*}
        \frac{1}{T} \sum_{t=1}^T f^{\rm reg}(\bw_t) - f^{\rm reg}_* \le \mathcal{O} \left( \frac{1}{\sqrt{T}} \right) ,
    \end{align*}
    where $f^{\rm reg}_*$ denotes the optimum value of $f^{\rm reg}$.
\end{theorem}

\paragraph{Memorization and Reasoning}

In our instruction following experiments in Appendix~\ref{appendix:stable_diffusion} and~\ref{app:ft}, we observe that LISA is much better than LoRA at memorization-centered tasks, such as Writing or depicting image details, while this gap is much smaller in reasoning-centered tasks like Code or Math. It is an intriguing observation since LISA emphasizes more on layer-wise width and restricts the depth of learned parameters, while LoRA focuses more on depth and restricts the representation space in each layer. It may suggest that width is crucial for memorization, while depth is important for reasoning, a similar phenomenon that echos the intuition of~\citep{cheng2016wide}. Based on the same intuition, it may be possible to combine the benefits of both and bring forth an even better PEFT method.

\section{Conclusion}
In this paper, we propose Layerwise Importance Sampled AdamW (LISA), an optimization algorithm that randomly freezes layers of LLM based on a given probability. Inspired by observations of LoRA's skewed weight norm distribution, a simple and memory-efficient freezing paradigm is introduced for LLM training. This paradigm achieves significant performance improvements over LoRA on downstream fine-tuning tasks with various models, including LLaMA-2-70B. Further experiments on domain-specific training also demonstrate its effectiveness, showing LISA's huge potential as a promising alternative to LoRA for LLM training.

\section*{Limitations}
\label{limit}
The major bottleneck of LISA is the same as LoRA, where during optimization, the forward pass still requires the model to be presented in the memory, leading to significant memory consumption. This limitation shall be compensated by approaches similar to QLoRA~\citep{dettmers2023qlora}, where we intend to conduct further experiments to verify its performance.

In addition, as suggested by the theoretical intuition, the strategy of E+H+2L in Section~\ref{sec:exp_moderate_ft} and E+H+4L in Section~\ref{sec:exp_large_ft} may not be the optimal importance sampling strategy, given it still sampled intermediate layers in a uniformly random fashion. We anticipate the optimizer's efficiency will be further improved when considering data sources and model architecture in the importance sampling procedure.

\bibliographystyle{acl_natbib}
\bibliography{ref}

\begin{thebibliography}{81}
\expandafter\ifx\csname natexlab\endcsname\relax\def\natexlab#1{#1}\fi

\bibitem[{Almazrouei et~al.(2023)Almazrouei, Alobeidli, Alshamsi, Cappelli, Cojocaru, Alhammadi, Daniele, Heslow, Launay, Malartic, Noune, Pannier, and Penedo}]{falcon180b}
Ebtesam Almazrouei, Hamza Alobeidli, Abdulaziz Alshamsi, Alessandro Cappelli, Ruxandra Cojocaru, Maitha Alhammadi, Mazzotta Daniele, Daniel Heslow, Julien Launay, Quentin Malartic, Badreddine Noune, Baptiste Pannier, and Guilherme Penedo. 2023.
\newblock The falcon series of language models: Towards open frontier models.

\bibitem[{Bengio et~al.(2007)Bengio, Lamblin, Popovici, and Larochelle}]{Bengio_Lamblin_Popovici_Larochelle_2007}
Yoshua Bengio, Pascal Lamblin, Dan Popovici, and Hugo Larochelle. 2007.
\newblock \href {https://doi.org/10.7551/mitpress/7503.003.0024} {\emph{Greedy Layer-Wise Training of Deep Networks}}, page 153–160.

\bibitem[{Biderman et~al.(2023)Biderman, Schoelkopf, Anthony, Bradley, O’Brien, Hallahan, Khan, Purohit, Prashanth, Raff et~al.}]{biderman2023pythia}
Stella Biderman, Hailey Schoelkopf, Quentin~Gregory Anthony, Herbie Bradley, Kyle O’Brien, Eric Hallahan, Mohammad~Aflah Khan, Shivanshu Purohit, USVSN~Sai Prashanth, Edward Raff, et~al. 2023.
\newblock Pythia: A suite for analyzing large language models across training and scaling.
\newblock In \emph{International Conference on Machine Learning}, pages 2397--2430. PMLR.

\bibitem[{Brock et~al.(2017)Brock, Lim, Ritchie, and Weston}]{brocker2017freezeout}
Andrew Brock, Theodore Lim, James~M. Ritchie, and Nick Weston. 2017.
\newblock \href {http://dblp.uni-trier.de/db/journals/corr/corr1706.html#BrockLRW17} {Freezeout: Accelerate training by progressively freezing layers.}
\newblock \emph{CoRR}, abs/1706.04983.

\bibitem[{Brown et~al.(2020)Brown, Mann, Ryder, Subbiah, Kaplan, Dhariwal, Neelakantan, Shyam, Sastry, Askell, Agarwal, Herbert-Voss, Krueger, Henighan, Child, Ramesh, Ziegler, Wu, Winter, Hesse, Chen, Sigler, Litwin, Gray, Chess, Clark, Berner, McCandlish, Radford, Sutskever, and Amodei}]{brown_language_2020}
Tom Brown, Benjamin Mann, Nick Ryder, Melanie Subbiah, Jared~D Kaplan, Prafulla Dhariwal, Arvind Neelakantan, Pranav Shyam, Girish Sastry, Amanda Askell, Sandhini Agarwal, Ariel Herbert-Voss, Gretchen Krueger, Tom Henighan, Rewon Child, Aditya Ramesh, Daniel Ziegler, Jeffrey Wu, Clemens Winter, Chris Hesse, Mark Chen, Eric Sigler, Mateusz Litwin, Scott Gray, Benjamin Chess, Jack Clark, Christopher Berner, Sam McCandlish, Alec Radford, Ilya Sutskever, and Dario Amodei. 2020.
\newblock \href {https://proceedings.neurips.cc/paper_files/paper/2020/file/1457c0d6bfcb4967418bfb8ac142f64a-Paper.pdf} {Language {Models} are {Few}-{Shot} {Learners}}.
\newblock In \emph{Advances in {Neural} {Information} {Processing} {Systems}}, volume~33, pages 1877--1901. Curran Associates, Inc.

\bibitem[{Chen et~al.(2016)Chen, Xu, Zhang, and Guestrin}]{chen2016training}
Tianqi Chen, Bing Xu, Chiyuan Zhang, and Carlos Guestrin. 2016.
\newblock Training deep nets with sublinear memory cost.
\newblock \emph{arXiv preprint arXiv:1604.06174}.

\bibitem[{Cheng et~al.(2016)Cheng, Koc, Harmsen, Shaked, Chandra, Aradhye, Anderson, Corrado, Chai, Ispir et~al.}]{cheng2016wide}
Heng-Tze Cheng, Levent Koc, Jeremiah Harmsen, Tal Shaked, Tushar Chandra, Hrishi Aradhye, Glen Anderson, Greg Corrado, Wei Chai, Mustafa Ispir, et~al. 2016.
\newblock Wide \& deep learning for recommender systems.
\newblock In \emph{Proceedings of the 1st workshop on deep learning for recommender systems}, pages 7--10.

\bibitem[{Chiang et~al.(2023)Chiang, Li, Lin, Sheng, Wu, Zhang, Zheng, Zhuang, Zhuang, Gonzalez, Stoica, and Xing}]{vicuna2023}
Wei-Lin Chiang, Zhuohan Li, Zi~Lin, Ying Sheng, Zhanghao Wu, Hao Zhang, Lianmin Zheng, Siyuan Zhuang, Yonghao Zhuang, Joseph~E. Gonzalez, Ion Stoica, and Eric~P. Xing. 2023.
\newblock \href {https://lmsys.org/blog/2023-03-30-vicuna/} {Vicuna: An open-source chatbot impressing gpt-4 with 90\%* chatgpt quality}.

\bibitem[{Chowdhery et~al.(2022)Chowdhery, Narang, Devlin, Bosma, Mishra, Roberts, Barham, Chung, Sutton, Gehrmann, Schuh, Shi, Tsvyashchenko, Maynez, Rao, Barnes, Tay, Shazeer, Prabhakaran, Reif, Du, Hutchinson, Pope, Bradbury, Austin, Isard, Gur-Ari, Yin, Duke, Levskaya, Ghemawat, Dev, Michalewski, Garc{\'i}a, Misra, Robinson, Fedus, Zhou, Ippolito, Luan, Lim, Zoph, Spiridonov, Sepassi, Dohan, Agrawal, Omernick, Dai, Pillai, Pellat, Lewkowycz, Moreira, Child, Polozov, Lee, Zhou, Wang, Saeta, D{\'i}az, Firat, Catasta, Wei, Meier-Hellstern, Eck, Dean, Petrov, and Fiedel}]{Chowdhery2022PaLMSL}
Aakanksha Chowdhery, Sharan Narang, Jacob Devlin, Maarten Bosma, Gaurav Mishra, Adam Roberts, Paul Barham, Hyung~Won Chung, Charles Sutton, Sebastian Gehrmann, Parker Schuh, Kensen Shi, Sasha Tsvyashchenko, Joshua Maynez, Abhishek Rao, Parker Barnes, Yi~Tay, Noam~M. Shazeer, Vinodkumar Prabhakaran, Emily Reif, Nan Du, Benton~C. Hutchinson, Reiner Pope, James Bradbury, Jacob Austin, Michael Isard, Guy Gur-Ari, Pengcheng Yin, Toju Duke, Anselm Levskaya, Sanjay Ghemawat, Sunipa Dev, Henryk Michalewski, Xavier Garc{\'i}a, Vedant Misra, Kevin Robinson, Liam Fedus, Denny Zhou, Daphne Ippolito, David Luan, Hyeontaek Lim, Barret Zoph, Alexander Spiridonov, Ryan Sepassi, David Dohan, Shivani Agrawal, Mark Omernick, Andrew~M. Dai, Thanumalayan~Sankaranarayana Pillai, Marie Pellat, Aitor Lewkowycz, Erica Moreira, Rewon Child, Oleksandr Polozov, Katherine Lee, Zongwei Zhou, Xuezhi Wang, Brennan Saeta, Mark D{\'i}az, Orhan Firat, Michele Catasta, Jason Wei, Kathleen~S. Meier-Hellstern, Douglas Eck, Jeff Dean, Slav Petrov,
  and Noah Fiedel. 2022.
\newblock Palm: Scaling language modeling with pathways.
\newblock \emph{J. Mach. Learn. Res.}, 24:240:1--240:113.

\bibitem[{Chuang et~al.(2023)Chuang, Xie, Luo, Kim, Glass, and He}]{chuang2023dola}
Yung-Sung Chuang, Yujia Xie, Hongyin Luo, Yoon Kim, James Glass, and Pengcheng He. 2023.
\newblock Dola: Decoding by contrasting layers improves factuality in large language models.
\newblock \emph{arXiv preprint arXiv:2309.03883}.

\bibitem[{Cobbe et~al.(2021)Cobbe, Kosaraju, Bavarian, Chen, Jun, Kaiser, Plappert, Tworek, Hilton, Nakano, Hesse, and Schulman}]{cobbe2021training}
Karl Cobbe, Vineet Kosaraju, Mohammad Bavarian, Mark Chen, Heewoo Jun, Lukasz Kaiser, Matthias Plappert, Jerry Tworek, Jacob Hilton, Reiichiro Nakano, Christopher Hesse, and John Schulman. 2021.
\newblock \href {http://arxiv.org/abs/2110.14168} {Training verifiers to solve math word problems}.

\bibitem[{Dao(2023)}]{dao2023flashattention2}
Tri Dao. 2023.
\newblock Flash{A}ttention-2: Faster attention with better parallelism and work partitioning.

\bibitem[{Dao et~al.(2022)Dao, Fu, Ermon, Rudra, and R{\'e}}]{dao2022flashattention}
Tri Dao, Daniel~Y. Fu, Stefano Ermon, Atri Rudra, and Christopher R{\'e}. 2022.
\newblock Flash{A}ttention: Fast and memory-efficient exact attention with {IO}-awareness.
\newblock In \emph{Advances in Neural Information Processing Systems}.

\bibitem[{Dettmers et~al.(2023)Dettmers, Pagnoni, Holtzman, and Zettlemoyer}]{dettmers2023qlora}
Tim Dettmers, Artidoro Pagnoni, Ari Holtzman, and Luke Zettlemoyer. 2023.
\newblock Qlora: Efficient finetuning of quantized llms.
\newblock \emph{arXiv preprint arXiv:2305.14314}.

\bibitem[{Devlin et~al.(2019)Devlin, Chang, Lee, and Toutanova}]{devlin-etal-2019-bert}
Jacob Devlin, Ming-Wei Chang, Kenton Lee, and Kristina Toutanova. 2019.
\newblock \href {https://doi.org/10.18653/v1/N19-1423} {{BERT}: Pre-training of deep bidirectional transformers for language understanding}.
\newblock In \emph{Proceedings of the 2019 Conference of the North {A}merican Chapter of the Association for Computational Linguistics: Human Language Technologies, Volume 1 (Long and Short Papers)}, pages 4171--4186, Minneapolis, Minnesota. Association for Computational Linguistics.

\bibitem[{Diao et~al.(2022)Diao, Huang, Xu, Li, Yong, Zhou, and Zhang}]{diao2022black}
Shizhe Diao, Zhichao Huang, Ruijia Xu, Xuechun Li, LIN Yong, Xiao Zhou, and Tong Zhang. 2022.
\newblock Black-box prompt learning for pre-trained language models.
\newblock \emph{Transactions on Machine Learning Research}.

\bibitem[{Diao et~al.(2023{\natexlab{a}})Diao, Pan, Dong, Shum, Zhang, Xiong, and Zhang}]{diao2023lmflow}
Shizhe Diao, Rui Pan, Hanze Dong, Ka~Shun Shum, Jipeng Zhang, Wei Xiong, and Tong Zhang. 2023{\natexlab{a}}.
\newblock Lmflow: An extensible toolkit for finetuning and inference of large foundation models.
\newblock \emph{arXiv preprint arXiv:2306.12420}.

\bibitem[{Diao et~al.(2023{\natexlab{b}})Diao, Wang, Lin, and Zhang}]{diao2023active}
Shizhe Diao, Pengcheng Wang, Yong Lin, and Tong Zhang. 2023{\natexlab{b}}.
\newblock \href {http://arxiv.org/abs/2302.12246} {Active prompting with chain-of-thought for large language models}.

\bibitem[{Diao et~al.(2021)Diao, Xu, Su, Jiang, Song, and Zhang}]{diao2021taming}
Shizhe Diao, Ruijia Xu, Hongjin Su, Yilei Jiang, Yan Song, and Tong Zhang. 2021.
\newblock Taming pre-trained language models with n-gram representations for low-resource domain adaptation.
\newblock In \emph{Proceedings of the 59th Annual Meeting of the Association for Computational Linguistics and the 11th International Joint Conference on Natural Language Processing (Volume 1: Long Papers)}, pages 3336--3349.

\bibitem[{Diao et~al.(2023{\natexlab{c}})Diao, Xu, Xu, Wang, and Zhang}]{diao2023mixture}
Shizhe Diao, Tianyang Xu, Ruijia Xu, Jiawei Wang, and Tong Zhang. 2023{\natexlab{c}}.
\newblock Mixture-of-domain-adapters: Decoupling and injecting domain knowledge to pre-trained language models memories.
\newblock \emph{arXiv preprint arXiv:2306.05406}.

\bibitem[{Ding et~al.(2022)Ding, Qin, Yang, Wei, Yang, Su, Hu, Chen, Chan, Chen et~al.}]{ding2022delta}
Ning Ding, Yujia Qin, Guang Yang, Fuchao Wei, Zonghan Yang, Yusheng Su, Shengding Hu, Yulin Chen, Chi-Min Chan, Weize Chen, et~al. 2022.
\newblock Delta tuning: A comprehensive study of parameter efficient methods for pre-trained language models.
\newblock \emph{arXiv preprint arXiv:2203.06904}.

\bibitem[{Fan et~al.(2024)Fan, Jiang, Li, Meng, Han, Shang, Sun, Wang, and Wang}]{fan2024not}
Siqi Fan, Xin Jiang, Xiang Li, Xuying Meng, Peng Han, Shuo Shang, Aixin Sun, Yequan Wang, and Zhongyuan Wang. 2024.
\newblock Not all layers of llms are necessary during inference.
\newblock \emph{arXiv preprint arXiv:2403.02181}.

\bibitem[{Gou et~al.(2023)Gou, Liu, Chen, Hong, Xu, Li, Yeung, Kwok, and Zhang}]{gou2023mixture}
Yunhao Gou, Zhili Liu, Kai Chen, Lanqing Hong, Hang Xu, Aoxue Li, Dit-Yan Yeung, James~T Kwok, and Yu~Zhang. 2023.
\newblock Mixture of cluster-conditional lora experts for vision-language instruction tuning.
\newblock \emph{arXiv preprint arXiv:2312.12379}.

\bibitem[{Hambardzumyan et~al.(2021)Hambardzumyan, Khachatrian, and May}]{hambardzumyan2021warp}
Karen Hambardzumyan, Hrant Khachatrian, and Jonathan May. 2021.
\newblock \href {https://doi.org/10.18653/v1/2021.acl-long.381} {{WARP}: {W}ord-level {A}dversarial {R}e{P}rogramming}.
\newblock In \emph{Proceedings of the 59th Annual Meeting of the Association for Computational Linguistics and the 11th International Joint Conference on Natural Language Processing (Volume 1: Long Papers)}, pages 4921--4933, Online. Association for Computational Linguistics.

\bibitem[{Han et~al.(2021)Han, Zhao, Ding, Liu, and Sun}]{han2021ptr}
Xu~Han, Weilin Zhao, Ning Ding, Zhiyuan Liu, and Maosong Sun. 2021.
\newblock \href {https://arxiv.org/abs/2105.11259} {{PTR: Prompt Tuning with Rules for Text Classification}}.
\newblock \emph{ArXiv preprint}, abs/2105.11259.

\bibitem[{Hendrycks et~al.(2020)Hendrycks, Burns, Basart, Zou, Mazeika, Song, and Steinhardt}]{hendrycks2020measuring}
Dan Hendrycks, Collin Burns, Steven Basart, Andy Zou, Mantas Mazeika, Dawn Song, and Jacob Steinhardt. 2020.
\newblock Measuring massive multitask language understanding.
\newblock \emph{arXiv preprint arXiv:2009.03300}.

\bibitem[{Hinton et~al.(2006)Hinton, Osindero, and Teh}]{Hinton_Osindero_Teh_2006}
Geoffrey~E. Hinton, Simon Osindero, and Yee-Whye Teh. 2006.
\newblock \href {https://doi.org/10.1162/neco.2006.18.7.1527} {A fast learning algorithm for deep belief nets}.
\newblock \emph{Neural Computation}, page 1527–1554.

\bibitem[{Ho et~al.(2020)Ho, Jain, and Abbeel}]{Ho2020DenoisingDP}
Jonathan Ho, Ajay Jain, and P.~Abbeel. 2020.
\newblock \href {https://api.semanticscholar.org/CorpusID:219955663} {Denoising diffusion probabilistic models}.
\newblock \emph{ArXiv}, abs/2006.11239.

\bibitem[{Houlsby et~al.(2019)Houlsby, Giurgiu, Jastrzebski, Morrone, De~Laroussilhe, Gesmundo, Attariyan, and Gelly}]{houlsby2019peft}
Neil Houlsby, Andrei Giurgiu, Stanislaw Jastrzebski, Bruna Morrone, Quentin De~Laroussilhe, Andrea Gesmundo, Mona Attariyan, and Sylvain Gelly. 2019.
\newblock Parameter-efficient transfer learning for nlp.
\newblock In \emph{International Conference on Machine Learning}, pages 2790--2799. PMLR.

\bibitem[{Hu et~al.(2022)Hu, yelong shen, Wallis, Allen-Zhu, Li, Wang, Wang, and Chen}]{hu2022lora}
Edward~J Hu, yelong shen, Phillip Wallis, Zeyuan Allen-Zhu, Yuanzhi Li, Shean Wang, Lu~Wang, and Weizhu Chen. 2022.
\newblock \href {https://openreview.net/forum?id=nZeVKeeFYf9} {Lo{RA}: Low-rank adaptation of large language models}.
\newblock In \emph{International Conference on Learning Representations}.

\bibitem[{Ji et~al.(2024)Ji, Liu, Zhang, Zhang, Zhao, Zhou, Zhang, Liu, and Zheng}]{ji2024advlora}
Yuheng Ji, Yue Liu, Zhicheng Zhang, Zhao Zhang, Yuting Zhao, Gang Zhou, Xingwei Zhang, Xinwang Liu, and Xiaolong Zheng. 2024.
\newblock \href {http://arxiv.org/abs/2404.13425} {Advlora: Adversarial low-rank adaptation of vision-language models}.

\bibitem[{Jiang et~al.(2023)Jiang, Sablayrolles, Mensch, Bamford, Chaplot, de~las Casas, Bressand, Lengyel, Lample, Saulnier, Lavaud, Lachaux, Stock, Scao, Lavril, Wang, Lacroix, and Sayed}]{jiang2023mistral}
Albert~Q. Jiang, Alexandre Sablayrolles, Arthur Mensch, Chris Bamford, Devendra~Singh Chaplot, Diego de~las Casas, Florian Bressand, Gianna Lengyel, Guillaume Lample, Lucile Saulnier, Lélio~Renard Lavaud, Marie-Anne Lachaux, Pierre Stock, Teven~Le Scao, Thibaut Lavril, Thomas Wang, Timothée Lacroix, and William~El Sayed. 2023.
\newblock \href {http://arxiv.org/abs/2310.06825} {Mistral 7b}.

\bibitem[{Jiang et~al.(2024)Jiang, Sablayrolles, Roux, Mensch, Savary, Bamford, Chaplot, Casas, Hanna, Bressand et~al.}]{jiang2024mixtral}
Albert~Q Jiang, Alexandre Sablayrolles, Antoine Roux, Arthur Mensch, Blanche Savary, Chris Bamford, Devendra~Singh Chaplot, Diego de~las Casas, Emma~Bou Hanna, Florian Bressand, et~al. 2024.
\newblock Mixtral of experts.
\newblock \emph{arXiv preprint arXiv:2401.04088}.

\bibitem[{Jin et~al.(2019)Jin, Dhingra, Liu, Cohen, and Lu}]{jin2019pubmedqa}
Qiao Jin, Bhuwan Dhingra, Zhengping Liu, William Cohen, and Xinghua Lu. 2019.
\newblock Pubmedqa: A dataset for biomedical research question answering.
\newblock In \emph{Proceedings of the 2019 Conference on Empirical Methods in Natural Language Processing and the 9th International Joint Conference on Natural Language Processing (EMNLP-IJCNLP)}, pages 2567--2577.

\bibitem[{Kingma and Ba(2014)}]{kingma2014adam}
Diederik~P Kingma and Jimmy Ba. 2014.
\newblock Adam: A method for stochastic optimization.
\newblock \emph{arXiv preprint arXiv:1412.6980}.

\bibitem[{Kloek and Van~Dijk(1978)}]{kloek1978bayesian}
Teun Kloek and Herman~K Van~Dijk. 1978.
\newblock Bayesian estimates of equation system parameters: an application of integration by monte carlo.
\newblock \emph{Econometrica: Journal of the Econometric Society}, pages 1--19.

\bibitem[{Li et~al.(2023)Li, Yuan, Dai, Zhang, Wang, and Tang}]{li2023smartfrz}
Sheng Li, Geng Yuan, Yue Dai, Youtao Zhang, Yanzhi Wang, and Xulong Tang. 2023.
\newblock \href {https://openreview.net/forum?id=i9UlAr1T_xl} {Smart{FRZ}: An efficient training framework using attention-based layer freezing}.
\newblock In \emph{The Eleventh International Conference on Learning Representations}.

\bibitem[{Li and Liang(2021)}]{li2021prefix}
Xiang~Lisa Li and Percy Liang. 2021.
\newblock Prefix-tuning: Optimizing continuous prompts for generation.
\newblock \emph{arXiv preprint arXiv:2101.00190}.

\bibitem[{Lialin et~al.(2023)Lialin, Shivagunde, Muckatira, and Rumshisky}]{lialin2023relora}
Vladislav Lialin, Namrata Shivagunde, Sherin Muckatira, and Anna Rumshisky. 2023.
\newblock \href {http://arxiv.org/abs/2307.05695} {Relora: High-rank training through low-rank updates}.

\bibitem[{Liu et~al.(2023)Liu, Li, Hall, Liang, and Ma}]{liu2023sophia}
Hong Liu, Zhiyuan Li, David Hall, Percy Liang, and Tengyu Ma. 2023.
\newblock Sophia: A scalable stochastic second-order optimizer for language model pre-training.
\newblock \emph{arXiv preprint arXiv:2305.14342}.

\bibitem[{Liu et~al.(2021{\natexlab{a}})Liu, Zheng, Du, Ding, Qian, Yang, and Tang}]{liu2021gpt}
Xiao Liu, Yanan Zheng, Zhengxiao Du, Ming Ding, Yujie Qian, Zhilin Yang, and Jie Tang. 2021{\natexlab{a}}.
\newblock Gpt understands, too.
\newblock \emph{arXiv preprint arXiv:2103.10385}.

\bibitem[{Liu et~al.(2021{\natexlab{b}})Liu, Agarwal, and Venkataraman}]{liu2021autofreeze}
Yuhan Liu, Saurabh Agarwal, and Shivaram Venkataraman. 2021{\natexlab{b}}.
\newblock \href {http://arxiv.org/abs/2102.01386} {Autofreeze: Automatically freezing model blocks to accelerate fine-tuning}.

\bibitem[{Loshchilov and Hutter(2017)}]{loshchilov2017adamw}
Ilya Loshchilov and Frank Hutter. 2017.
\newblock Decoupled weight decay regularization.
\newblock \emph{arXiv preprint arXiv:1711.05101}.

\bibitem[{Luo et~al.(2024)Luo, Yu, and Li}]{luo2024badam}
Qijun Luo, Hengxu Yu, and Xiao Li. 2024.
\newblock \href {http://arxiv.org/abs/2404.02827} {Badam: A memory efficient full parameter training method for large language models}.

\bibitem[{Luo et~al.(2023)Luo, Tan, Huang, Li, and Zhao}]{luo2023latent}
Simian Luo, Yiqin Tan, Longbo Huang, Jian Li, and Hang Zhao. 2023.
\newblock Latent consistency models: Synthesizing high-resolution images with few-step inference.
\newblock \emph{arXiv preprint arXiv:2310.04378}.

\bibitem[{Malladi et~al.(2023)Malladi, Gao, Nichani, Damian, Lee, Chen, and Arora}]{malladi2023finetuning}
Sadhika Malladi, Tianyu Gao, Eshaan Nichani, Alex Damian, Jason~D. Lee, Danqi Chen, and Sanjeev Arora. 2023.
\newblock \href {https://openreview.net/forum?id=Vota6rFhBQ} {Fine-tuning language models with just forward passes}.
\newblock In \emph{Thirty-seventh Conference on Neural Information Processing Systems}.

\bibitem[{Meng et~al.(2024)Meng, Wang, and Zhang}]{meng2024pissa}
Fanxu Meng, Zhaohui Wang, and Muhan Zhang. 2024.
\newblock \href {http://arxiv.org/abs/2404.02948} {Pissa: Principal singular values and singular vectors adaptation of large language models}.

\bibitem[{OpenAI et~al.(2023)OpenAI, Achiam, Adler, Agarwal, Ahmad, Akkaya, Aleman, Almeida, Altenschmidt, Altman, Anadkat, Avila, Babuschkin, Balaji, Balcom, Baltescu, Bao, Bavarian, Belgum, Bello, Berdine, Bernadett-Shapiro, Berner, Bogdonoff, Boiko, Boyd, Brakman, Brockman, Brooks, Brundage, Button, Cai, Campbell, Cann, Carey, Carlson, Carmichael, Chan, Chang, Chantzis, Chen, Chen, Chen, Chen, Chen, Chess, Cho, Chu, Chung, Cummings, Currier, Dai, Decareaux, Degry, Deutsch, Deville, Dhar, Dohan, Dowling, Dunning, Ecoffet, Eleti, Eloundou, Farhi, Fedus, Felix, Fishman, Forte, Fulford, Gao, Georges, Gibson, Goel, Gogineni, Goh, Gontijo-Lopes, Gordon, Grafstein, Gray, Greene, Gross, Gu, Guo, Hallacy, Han, Harris, He, Heaton, Heidecke, Hesse, Hickey, Hickey, Hoeschele, Houghton, Hsu, Hu, Hu, Huizinga, Jain, Jain, Jang, Jiang, Jiang, Jin, Jin, Jomoto, Jonn, Jun, Kaftan, Łukasz Kaiser, Kamali, Kanitscheider, Keskar, Khan, Kilpatrick, Kim, Kim, Kim, Kirchner, Kiros, Knight, Kokotajlo, Łukasz Kondraciuk,
  Kondrich, Konstantinidis, Kosic, Krueger, Kuo, Lampe, Lan, Lee, Leike, Leung, Levy, Li, Lim, Lin, Lin, Litwin, Lopez, Lowe, Lue, Makanju, Malfacini, Manning, Markov, Markovski, Martin, Mayer, Mayne, McGrew, McKinney, McLeavey, McMillan, McNeil, Medina, Mehta, Menick, Metz, Mishchenko, Mishkin, Monaco, Morikawa, Mossing, Mu, Murati, Murk, Mély, Nair, Nakano, Nayak, Neelakantan, Ngo, Noh, Ouyang, O'Keefe, Pachocki, Paino, Palermo, Pantuliano, Parascandolo, Parish, Parparita, Passos, Pavlov, Peng, Perelman, de~Avila Belbute~Peres, Petrov, de~Oliveira~Pinto, Michael, Pokorny, Pokrass, Pong, Powell, Power, Power, Proehl, Puri, Radford, Rae, Ramesh, Raymond, Real, Rimbach, Ross, Rotsted, Roussez, Ryder, Saltarelli, Sanders, Santurkar, Sastry, Schmidt, Schnurr, Schulman, Selsam, Sheppard, Sherbakov, Shieh, Shoker, Shyam, Sidor, Sigler, Simens, Sitkin, Slama, Sohl, Sokolowsky, Song, Staudacher, Such, Summers, Sutskever, Tang, Tezak, Thompson, Tillet, Tootoonchian, Tseng, Tuggle, Turley, Tworek, Uribe, Vallone,
  Vijayvergiya, Voss, Wainwright, Wang, Wang, Wang, Ward, Wei, Weinmann, Welihinda, Welinder, Weng, Weng, Wiethoff, Willner, Winter, Wolrich, Wong, Workman, Wu, Wu, Wu, Xiao, Xu, Yoo, Yu, Yuan, Zaremba, Zellers, Zhang, Zhang, Zhao, Zheng, Zhuang, Zhuk, and Zoph}]{openai2023gpt4}
OpenAI, Josh Achiam, Steven Adler, Sandhini Agarwal, Lama Ahmad, Ilge Akkaya, Florencia~Leoni Aleman, Diogo Almeida, Janko Altenschmidt, Sam Altman, Shyamal Anadkat, Red Avila, Igor Babuschkin, Suchir Balaji, Valerie Balcom, Paul Baltescu, Haiming Bao, Mo~Bavarian, Jeff Belgum, Irwan Bello, Jake Berdine, Gabriel Bernadett-Shapiro, Christopher Berner, Lenny Bogdonoff, Oleg Boiko, Madelaine Boyd, Anna-Luisa Brakman, Greg Brockman, Tim Brooks, Miles Brundage, Kevin Button, Trevor Cai, Rosie Campbell, Andrew Cann, Brittany Carey, Chelsea Carlson, Rory Carmichael, Brooke Chan, Che Chang, Fotis Chantzis, Derek Chen, Sully Chen, Ruby Chen, Jason Chen, Mark Chen, Ben Chess, Chester Cho, Casey Chu, Hyung~Won Chung, Dave Cummings, Jeremiah Currier, Yunxing Dai, Cory Decareaux, Thomas Degry, Noah Deutsch, Damien Deville, Arka Dhar, David Dohan, Steve Dowling, Sheila Dunning, Adrien Ecoffet, Atty Eleti, Tyna Eloundou, David Farhi, Liam Fedus, Niko Felix, Simón~Posada Fishman, Juston Forte, Isabella Fulford, Leo Gao,
  Elie Georges, Christian Gibson, Vik Goel, Tarun Gogineni, Gabriel Goh, Rapha Gontijo-Lopes, Jonathan Gordon, Morgan Grafstein, Scott Gray, Ryan Greene, Joshua Gross, Shixiang~Shane Gu, Yufei Guo, Chris Hallacy, Jesse Han, Jeff Harris, Yuchen He, Mike Heaton, Johannes Heidecke, Chris Hesse, Alan Hickey, Wade Hickey, Peter Hoeschele, Brandon Houghton, Kenny Hsu, Shengli Hu, Xin Hu, Joost Huizinga, Shantanu Jain, Shawn Jain, Joanne Jang, Angela Jiang, Roger Jiang, Haozhun Jin, Denny Jin, Shino Jomoto, Billie Jonn, Heewoo Jun, Tomer Kaftan, Łukasz Kaiser, Ali Kamali, Ingmar Kanitscheider, Nitish~Shirish Keskar, Tabarak Khan, Logan Kilpatrick, Jong~Wook Kim, Christina Kim, Yongjik Kim, Hendrik Kirchner, Jamie Kiros, Matt Knight, Daniel Kokotajlo, Łukasz Kondraciuk, Andrew Kondrich, Aris Konstantinidis, Kyle Kosic, Gretchen Krueger, Vishal Kuo, Michael Lampe, Ikai Lan, Teddy Lee, Jan Leike, Jade Leung, Daniel Levy, Chak~Ming Li, Rachel Lim, Molly Lin, Stephanie Lin, Mateusz Litwin, Theresa Lopez, Ryan Lowe,
  Patricia Lue, Anna Makanju, Kim Malfacini, Sam Manning, Todor Markov, Yaniv Markovski, Bianca Martin, Katie Mayer, Andrew Mayne, Bob McGrew, Scott~Mayer McKinney, Christine McLeavey, Paul McMillan, Jake McNeil, David Medina, Aalok Mehta, Jacob Menick, Luke Metz, Andrey Mishchenko, Pamela Mishkin, Vinnie Monaco, Evan Morikawa, Daniel Mossing, Tong Mu, Mira Murati, Oleg Murk, David Mély, Ashvin Nair, Reiichiro Nakano, Rajeev Nayak, Arvind Neelakantan, Richard Ngo, Hyeonwoo Noh, Long Ouyang, Cullen O'Keefe, Jakub Pachocki, Alex Paino, Joe Palermo, Ashley Pantuliano, Giambattista Parascandolo, Joel Parish, Emy Parparita, Alex Passos, Mikhail Pavlov, Andrew Peng, Adam Perelman, Filipe de~Avila Belbute~Peres, Michael Petrov, Henrique~Ponde de~Oliveira~Pinto, Michael, Pokorny, Michelle Pokrass, Vitchyr Pong, Tolly Powell, Alethea Power, Boris Power, Elizabeth Proehl, Raul Puri, Alec Radford, Jack Rae, Aditya Ramesh, Cameron Raymond, Francis Real, Kendra Rimbach, Carl Ross, Bob Rotsted, Henri Roussez, Nick Ryder,
  Mario Saltarelli, Ted Sanders, Shibani Santurkar, Girish Sastry, Heather Schmidt, David Schnurr, John Schulman, Daniel Selsam, Kyla Sheppard, Toki Sherbakov, Jessica Shieh, Sarah Shoker, Pranav Shyam, Szymon Sidor, Eric Sigler, Maddie Simens, Jordan Sitkin, Katarina Slama, Ian Sohl, Benjamin Sokolowsky, Yang Song, Natalie Staudacher, Felipe~Petroski Such, Natalie Summers, Ilya Sutskever, Jie Tang, Nikolas Tezak, Madeleine Thompson, Phil Tillet, Amin Tootoonchian, Elizabeth Tseng, Preston Tuggle, Nick Turley, Jerry Tworek, Juan Felipe~Cerón Uribe, Andrea Vallone, Arun Vijayvergiya, Chelsea Voss, Carroll Wainwright, Justin~Jay Wang, Alvin Wang, Ben Wang, Jonathan Ward, Jason Wei, CJ~Weinmann, Akila Welihinda, Peter Welinder, Jiayi Weng, Lilian Weng, Matt Wiethoff, Dave Willner, Clemens Winter, Samuel Wolrich, Hannah Wong, Lauren Workman, Sherwin Wu, Jeff Wu, Michael Wu, Kai Xiao, Tao Xu, Sarah Yoo, Kevin Yu, Qiming Yuan, Wojciech Zaremba, Rowan Zellers, Chong Zhang, Marvin Zhang, Shengjia Zhao, Tianhao
  Zheng, Juntang Zhuang, William Zhuk, and Barret Zoph. 2023.
\newblock \href {http://arxiv.org/abs/2303.08774} {Gpt-4 technical report}.

\bibitem[{Ouyang et~al.(2022)Ouyang, Wu, Jiang, Almeida, Wainwright, Mishkin, Zhang, Agarwal, Slama, Ray, Schulman, Hilton, Kelton, Miller, Simens, Askell, Welinder, Christiano, Leike, and Lowe}]{ouyang2022instructgpt}
Long Ouyang, Jeffrey Wu, Xu~Jiang, Diogo Almeida, Carroll Wainwright, Pamela Mishkin, Chong Zhang, Sandhini Agarwal, Katarina Slama, Alex Ray, John Schulman, Jacob Hilton, Fraser Kelton, Luke Miller, Maddie Simens, Amanda Askell, Peter Welinder, Paul~F Christiano, Jan Leike, and Ryan Lowe. 2022.
\newblock Training language models to follow instructions with human feedback.
\newblock In \emph{Advances in Neural Information Processing Systems}, volume~35, pages 27730--27744. Curran Associates, Inc.

\bibitem[{Paster et~al.(2023)Paster, Santos, Azerbayev, and Ba}]{paster2023openwebmath}
Keiran Paster, Marco~Dos Santos, Zhangir Azerbayev, and Jimmy Ba. 2023.
\newblock \href {http://arxiv.org/abs/2310.06786} {Openwebmath: An open dataset of high-quality mathematical web text}.

\bibitem[{Paszke et~al.(2019)Paszke, Gross, Massa, Lerer, Bradbury, Chanan, Killeen, Lin, Gimelshein, Antiga et~al.}]{paszke2019pytorch}
Adam Paszke, Sam Gross, Francisco Massa, Adam Lerer, James Bradbury, Gregory Chanan, Trevor Killeen, Zeming Lin, Natalia Gimelshein, Luca Antiga, et~al. 2019.
\newblock Pytorch: An imperative style, high-performance deep learning library.
\newblock \emph{Advances in neural information processing systems}, 32.

\bibitem[{Peng et~al.(2023)Peng, Li, He, Galley, and Gao}]{peng2023instruction}
Baolin Peng, Chunyuan Li, Pengcheng He, Michel Galley, and Jianfeng Gao. 2023.
\newblock Instruction tuning with gpt-4.
\newblock \emph{arXiv preprint arXiv:2304.03277}.

\bibitem[{Qin and Eisner(2021)}]{qin2021learning}
Guanghui Qin and Jason Eisner. 2021.
\newblock \href {https://doi.org/10.18653/v1/2021.naacl-main.410} {Learning how to ask: Querying {LM}s with mixtures of soft prompts}.
\newblock In \emph{Proceedings of the 2021 Conference of the North American Chapter of the Association for Computational Linguistics: Human Language Technologies}, pages 5203--5212, Online. Association for Computational Linguistics.

\bibitem[{Radford et~al.(2019)Radford, Wu, Child, Luan, Amodei, and Sutskever}]{radford2019gpt2}
Alec Radford, Jeff Wu, Rewon Child, David Luan, Dario Amodei, and Ilya Sutskever. 2019.
\newblock Language models are unsupervised multitask learners.

\bibitem[{Raffel et~al.(2020)Raffel, Shazeer, Roberts, Lee, Narang, Matena, Zhou, Li, and Liu}]{raffel2020t5}
Colin Raffel, Noam Shazeer, Adam Roberts, Katherine Lee, Sharan Narang, Michael Matena, Yanqi Zhou, Wei Li, and Peter~J. Liu. 2020.
\newblock Exploring the limits of transfer learning with a unified text-to-text transformer.
\newblock \emph{J. Mach. Learn. Res.}, 21(1).

\bibitem[{Rajbhandari et~al.(2020)Rajbhandari, Rasley, Ruwase, and He}]{rajbhandari2020zero}
Samyam Rajbhandari, Jeff Rasley, Olatunji Ruwase, and Yuxiong He. 2020.
\newblock Zero: Memory optimizations toward training trillion parameter models.
\newblock In \emph{SC20: International Conference for High Performance Computing, Networking, Storage and Analysis}, pages 1--16. IEEE.

\bibitem[{Rasley et~al.(2020)Rasley, Rajbhandari, Ruwase, and He}]{rasley2020deepspeed}
Jeff Rasley, Samyam Rajbhandari, Olatunji Ruwase, and Yuxiong He. 2020.
\newblock Deepspeed: System optimizations enable training deep learning models with over 100 billion parameters.
\newblock In \emph{Proceedings of the 26th ACM SIGKDD International Conference on Knowledge Discovery \& Data Mining}, pages 3505--3506.

\bibitem[{Reddi et~al.(2019)Reddi, Kale, and Kumar}]{reddi2019amsgrad}
Sashank~J Reddi, Satyen Kale, and Sanjiv Kumar. 2019.
\newblock On the convergence of adam and beyond.
\newblock \emph{arXiv preprint arXiv:1904.09237}.

\bibitem[{Ren et~al.(2021)Ren, Rajbhandari, Aminabadi, Ruwase, Yang, Zhang, Li, and He}]{ren2021zerooffload}
Jie Ren, Samyam Rajbhandari, Reza~Yazdani Aminabadi, Olatunji Ruwase, Shuangyan Yang, Minjia Zhang, Dong Li, and Yuxiong He. 2021.
\newblock \href {http://arxiv.org/abs/2101.06840} {Zero-offload: Democratizing billion-scale model training}.

\bibitem[{Rozière et~al.(2023)Rozière, Gehring, Gloeckle, Sootla, Gat, Tan, Adi, Liu, Remez, Rapin, Kozhevnikov, Evtimov, Bitton, Bhatt, Ferrer, Grattafiori, Xiong, Défossez, Copet, Azhar, Touvron, Martin, Usunier, Scialom, and Synnaeve}]{rozière2023code}
Baptiste Rozière, Jonas Gehring, Fabian Gloeckle, Sten Sootla, Itai Gat, Xiaoqing~Ellen Tan, Yossi Adi, Jingyu Liu, Tal Remez, Jérémy Rapin, Artyom Kozhevnikov, Ivan Evtimov, Joanna Bitton, Manish Bhatt, Cristian~Canton Ferrer, Aaron Grattafiori, Wenhan Xiong, Alexandre Défossez, Jade Copet, Faisal Azhar, Hugo Touvron, Louis Martin, Nicolas Usunier, Thomas Scialom, and Gabriel Synnaeve. 2023.
\newblock \href {http://arxiv.org/abs/2308.12950} {Code llama: Open foundation models for code}.

\bibitem[{Sakaguchi et~al.(2021)Sakaguchi, Bras, Bhagavatula, and Choi}]{sakaguchi2021winogrande}
Keisuke Sakaguchi, Ronan~Le Bras, Chandra Bhagavatula, and Yejin Choi. 2021.
\newblock Winogrande: An adversarial winograd schema challenge at scale.
\newblock \emph{Communications of the ACM}, 64(9):99--106.

\bibitem[{Scao et~al.(2022)Scao, Fan, Akiki, Pavlick, Ili{\'c}, Hesslow, Castagn{\'e}, Luccioni, Yvon et~al.}]{workshop2022bloom}
Teven~Le Scao, Angela Fan, Christopher Akiki, Ellie Pavlick, Suzana Ili{\'c}, Daniel Hesslow, Roman Castagn{\'e}, Alexandra~Sasha Luccioni, Fran{\c{c}}ois Yvon, et~al. 2022.
\newblock Bloom: A 176b-parameter open-access multilingual language model.
\newblock \emph{arXiv preprint arXiv:2211.05100}.

\bibitem[{Shum et~al.(2023)Shum, Diao, and Zhang}]{shum2023automatic}
Kashun Shum, Shizhe Diao, and Tong Zhang. 2023.
\newblock Automatic prompt augmentation and selection with chain-of-thought from labeled data.
\newblock In \emph{Findings of the Association for Computational Linguistics: EMNLP 2023}, pages 12113--12139.

\bibitem[{SONG et~al.(2024)SONG, Zhao, Majumder, and Lin}]{song2024increasing}
Haobo SONG, Hao Zhao, Soumajit Majumder, and Tao Lin. 2024.
\newblock \href {https://openreview.net/forum?id=H3IUunLy8s} {Increasing model capacity for free: A simple strategy for parameter efficient fine-tuning}.
\newblock In \emph{The Twelfth International Conference on Learning Representations}.

\bibitem[{Srivastava et~al.(2014)Srivastava, Hinton, Krizhevsky, Sutskever, and Salakhutdinov}]{JMLR:v15:srivastava14a:dropout}
Nitish Srivastava, Geoffrey Hinton, Alex Krizhevsky, Ilya Sutskever, and Ruslan Salakhutdinov. 2014.
\newblock \href {http://jmlr.org/papers/v15/srivastava14a.html} {Dropout: A simple way to prevent neural networks from overfitting}.
\newblock \emph{Journal of Machine Learning Research}, 15(56):1929--1958.

\bibitem[{Taori et~al.(2023)Taori, Gulrajani, Zhang, Dubois, Li, Guestrin, Liang, and Hashimoto}]{alpaca}
Rohan Taori, Ishaan Gulrajani, Tianyi Zhang, Yann Dubois, Xuechen Li, Carlos Guestrin, Percy Liang, and Tatsunori~B. Hashimoto. 2023.
\newblock Stanford alpaca: An instruction-following llama model.
\newblock \url{https://github.com/tatsu-lab/stanford_alpaca}.

\bibitem[{Touvron et~al.(2023{\natexlab{a}})Touvron, Lavril, Izacard, Martinet, Lachaux, Lacroix, Rozi{\`e}re, Goyal, Hambro, Azhar et~al.}]{touvron2023llama}
Hugo Touvron, Thibaut Lavril, Gautier Izacard, Xavier Martinet, Marie-Anne Lachaux, Timoth{\'e}e Lacroix, Baptiste Rozi{\`e}re, Naman Goyal, Eric Hambro, Faisal Azhar, et~al. 2023{\natexlab{a}}.
\newblock Llama: Open and efficient foundation language models.
\newblock \emph{arXiv preprint arXiv:2302.13971}.

\bibitem[{Touvron et~al.(2023{\natexlab{b}})Touvron, Martin, Stone, Albert, Almahairi, Babaei, Bashlykov, Batra, Bhargava, Bhosale et~al.}]{touvron2023llama2}
Hugo Touvron, Louis Martin, Kevin Stone, Peter Albert, Amjad Almahairi, Yasmine Babaei, Nikolay Bashlykov, Soumya Batra, Prajjwal Bhargava, Shruti Bhosale, et~al. 2023{\natexlab{b}}.
\newblock Llama 2: Open foundation and fine-tuned chat models.
\newblock \emph{arXiv preprint arXiv:2307.09288}.

\bibitem[{Vaswani et~al.(2017)Vaswani, Shazeer, Parmar, Uszkoreit, Jones, Gomez, Kaiser, and Polosukhin}]{Vaswani_Shazeer_Parmar_Uszkoreit_Jones_Gomez_Kaiser_Polosukhin_2017}
Ashish Vaswani, Noam Shazeer, Niki Parmar, Jakob Uszkoreit, Llion Jones, AidanN. Gomez, Lukasz Kaiser, and Illia Polosukhin. 2017.
\newblock Attention is all you need.
\newblock \emph{Neural Information Processing Systems,Neural Information Processing Systems}.

\bibitem[{Wangni et~al.(2018)Wangni, Wang, Liu, and Zhang}]{wangni2018gradient}
Jianqiao Wangni, Jialei Wang, Ji~Liu, and Tong Zhang. 2018.
\newblock Gradient sparsification for communication-efficient distributed optimization.
\newblock \emph{Advances in Neural Information Processing Systems}, 31.

\bibitem[{Wei et~al.(2022)Wei, Wang, Schuurmans, Bosma, Xia, Chi, Le, Zhou et~al.}]{wei2022chain}
Jason Wei, Xuezhi Wang, Dale Schuurmans, Maarten Bosma, Fei Xia, Ed~Chi, Quoc~V Le, Denny Zhou, et~al. 2022.
\newblock Chain-of-thought prompting elicits reasoning in large language models.
\newblock \emph{Advances in Neural Information Processing Systems}, 35:24824--24837.

\bibitem[{You et~al.(2017)You, Gitman, and Ginsburg}]{you2017lars}
Yang You, Igor Gitman, and Boris Ginsburg. 2017.
\newblock Large batch training of convolutional networks.
\newblock \emph{arXiv preprint arXiv:1708.03888}.

\bibitem[{You et~al.(2019)You, Li, Reddi, Hseu, Kumar, Bhojanapalli, Song, Demmel, Keutzer, and Hsieh}]{you2019lamb}
Yang You, Jing Li, Sashank Reddi, Jonathan Hseu, Sanjiv Kumar, Srinadh Bhojanapalli, Xiaodan Song, James Demmel, Kurt Keutzer, and Cho-Jui Hsieh. 2019.
\newblock Large batch optimization for deep learning: Training bert in 76 minutes.
\newblock \emph{arXiv preprint arXiv:1904.00962}.

\bibitem[{Zhang et~al.(2024)Zhang, Zeng, Wang, and Lu}]{zhang2024tinyllama}
Peiyuan Zhang, Guangtao Zeng, Tianduo Wang, and Wei Lu. 2024.
\newblock \href {http://arxiv.org/abs/2401.02385} {Tinyllama: An open-source small language model}.

\bibitem[{Zhang et~al.(2022)Zhang, Roller, Goyal, Artetxe, Chen, Chen, Dewan, Diab, Li, Lin et~al.}]{zhang2022opt}
Susan Zhang, Stephen Roller, Naman Goyal, Mikel Artetxe, Moya Chen, Shuohui Chen, Christopher Dewan, Mona Diab, Xian Li, Xi~Victoria Lin, et~al. 2022.
\newblock Opt: Open pre-trained transformer language models.
\newblock \emph{arXiv preprint arXiv:2205.01068}.

\bibitem[{Zhao et~al.(2024)Zhao, Zhang, Chen, Wang, Anandkumar, and Tian}]{zhao2024galore}
Jiawei Zhao, Zhenyu Zhang, Beidi Chen, Zhangyang Wang, Anima Anandkumar, and Yuandong Tian. 2024.
\newblock Galore: Memory-efficient llm training by gradient low-rank projection.
\newblock \emph{arXiv preprint arXiv:2403.03507}.

\bibitem[{Zhao and Zhang(2015)}]{zhao2015stochastic}
Peilin Zhao and Tong Zhang. 2015.
\newblock Stochastic optimization with importance sampling for regularized loss minimization.
\newblock In \emph{international conference on machine learning}, pages 1--9. PMLR.

\bibitem[{Zheng et~al.(2023)Zheng, Chiang, Sheng, Zhuang, Wu, Zhuang, Lin, Li, Li, Xing, Zhang, Gonzalez, and Stoica}]{zheng2023judging}
Lianmin Zheng, Wei-Lin Chiang, Ying Sheng, Siyuan Zhuang, Zhanghao Wu, Yonghao Zhuang, Zi~Lin, Zhuohan Li, Dacheng Li, Eric.~P Xing, Hao Zhang, Joseph~E. Gonzalez, and Ion Stoica. 2023.
\newblock \href {http://arxiv.org/abs/2306.05685} {Judging llm-as-a-judge with mt-bench and chatbot arena}.

\bibitem[{Zhong et~al.(2023)Zhong, Cui, Guo, Liang, Lu, Wang, Saied, Chen, and Duan}]{zhong2023agieval}
Wanjun Zhong, Ruixiang Cui, Yiduo Guo, Yaobo Liang, Shuai Lu, Yanlin Wang, Amin Saied, Weizhu Chen, and Nan Duan. 2023.
\newblock \href {http://arxiv.org/abs/2304.06364} {Agieval: A human-centric benchmark for evaluating foundation models}.

\bibitem[{Zhong et~al.(2021)Zhong, Friedman, and Chen}]{zhong2021factual}
Zexuan Zhong, Dan Friedman, and Danqi Chen. 2021.
\newblock \href {https://doi.org/10.18653/v1/2021.naacl-main.398} {Factual probing is [{MASK}]: Learning vs. learning to recall}.
\newblock In \emph{Proceedings of the 2021 Conference of the North American Chapter of the Association for Computational Linguistics: Human Language Technologies}, pages 5017--5033, Online. Association for Computational Linguistics.

\bibitem[{Zhou et~al.(2020)Zhou, Zhang, Zhu, Zheng, and Wu}]{zhou2020randomized}
Yangfan Zhou, Mingchuan Zhang, Junlong Zhu, Ruijuan Zheng, and Qingtao Wu. 2020.
\newblock A randomized block-coordinate adam online learning optimization algorithm.
\newblock \emph{Neural Computing and Applications}, 32(16):12671--12684.

\end{thebibliography}

\newpage
\appendix

\section{Additional Experiments}
\label{appendix:extra_exps}

\subsection{Image Generation}
\label{appendix:stable_diffusion}

Stable Diffusion has emerged as a powerful approach for generating high-quality images by leveraging the diffusion process in a latent space~\citep{Ho2020DenoisingDP}. This method involves diffusing the data distribution over time and iteratively refining the generated samples to enhance their realism and fidelity. The key innovation lies in operating within the latent space of a pre-trained autoencoder, significantly reducing computational complexity while maintaining high-quality outputs. The Latent Consistency Model (LCM) further improves the efficiency and performance of diffusion models in the latent space by incorporating a consistency loss, which penalizes deviations from the expected latent trajectories~\citep{luo2023latent}. This ensures that the generated samples remain coherent and visually plausible throughout the diffusion steps.

\begin{figure}[!htbp]
\vspace{-1.2cm}
\centering
\hspace*{-2.5cm}
\includegraphics[width=1.3\linewidth]{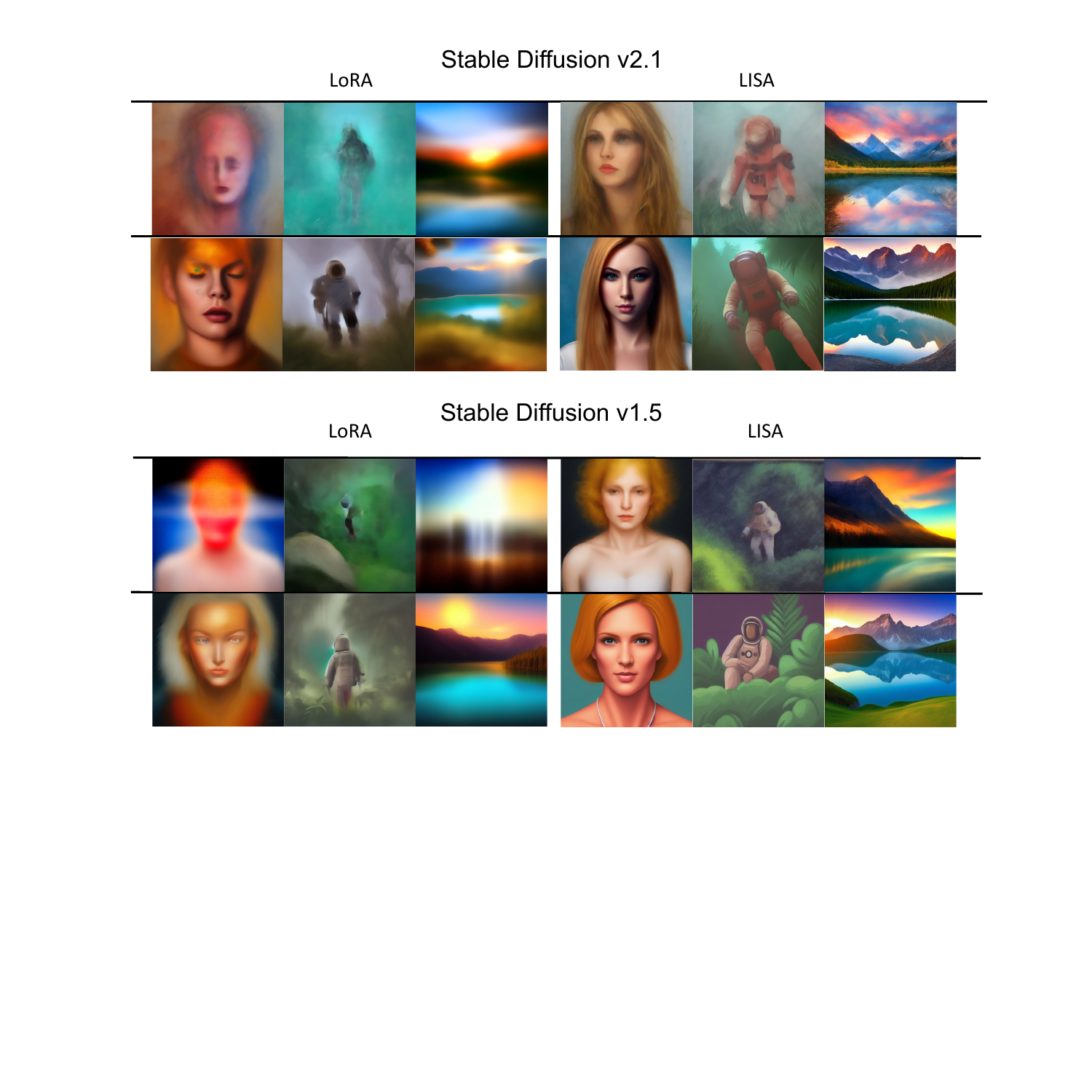}
\vspace{-7cm}
\caption{Generated images using LoRA (left) and LISA (right) on Stable Diffusion v2.1 model and Stable Diffusion v1.5. \textbf{First row}: number of inference step = 2. \textbf{Second row}: number of inference step = 10.}
\label{fig:sd_compare_v2_1}
\end{figure}

In our experiments, we evaluated the performance of LCM against other fine-tuning methods such as LoRA, and LISA using latent consistency distillation to distill stable diffusion v1.5 and v2.1 model. We utilized the official code\footnote{https://github.com/huggingface/diffusers/tree/main/examples/consistency\_distillation} and parameters for both LISA and LoRA from the Diffusers library to ensure consistency and comparability of results. Adjustments were made only to the batch size and accumulation steps to achieve a balanced trade-off between computational efficiency and performance. Figure \ref{fig:sd_compare_v2_1} is the generated image comparison. Observations reveal that LISA can generate higher-quality images in fewer inference steps. The images generated using LISA display more intricate details and sharper clarity, particularly evident in the distinct facial features and environment textures. In contrast, the LoRA-generated images offer a softer, more blended aesthetic with a dream-like quality, emphasizing smooth transitions over precise detail. The prompts for images in figure \ref{fig:sd_compare_v2_1} in the left-to-right are given below
\begin{itemize}
    \item Self-portrait oil painting, a beautiful cyborg with golden hair, 8k.
    \item Astronaut in a jungle, cold color palette, muted colors, detailed, 8k.
    \item A photo of a beautiful mountain with a realistic sunset and blue lake, a highly detailed masterpiece.
\end{itemize}

\subsection{Instruction Following Fine-tuning}
\label{app:ft}

Table~\ref{tab:all_mt_bench_full} offers a comprehensive evaluation of three fine-tuning methods—Full Parameter Fine-Tuning (FT), Low-Rank Adaptation (LoRA), Gradient Low-Rank Projection(GaLore), and Layerwise Importance Sampling AdamW (LISA)—across a diverse set of tasks including Writing, Roleplay, Reasoning, Math, Extraction, STEM, and Humanities within the MT-Bench benchmark. The results demonstrate LISA's superior performance, which surpasses LoRA, GaLore, and full parameter tuning in most settings. Notably, LISA consistently outperforms LoRA and full parameter tuning in domains such as Writing, STEM, and Humanities. This implies that LISA can benefit memorization tasks, while LoRA partially favors reasoning tasks.

\begin{figure}
\begin{center}

\includegraphics[width=0.45\linewidth]{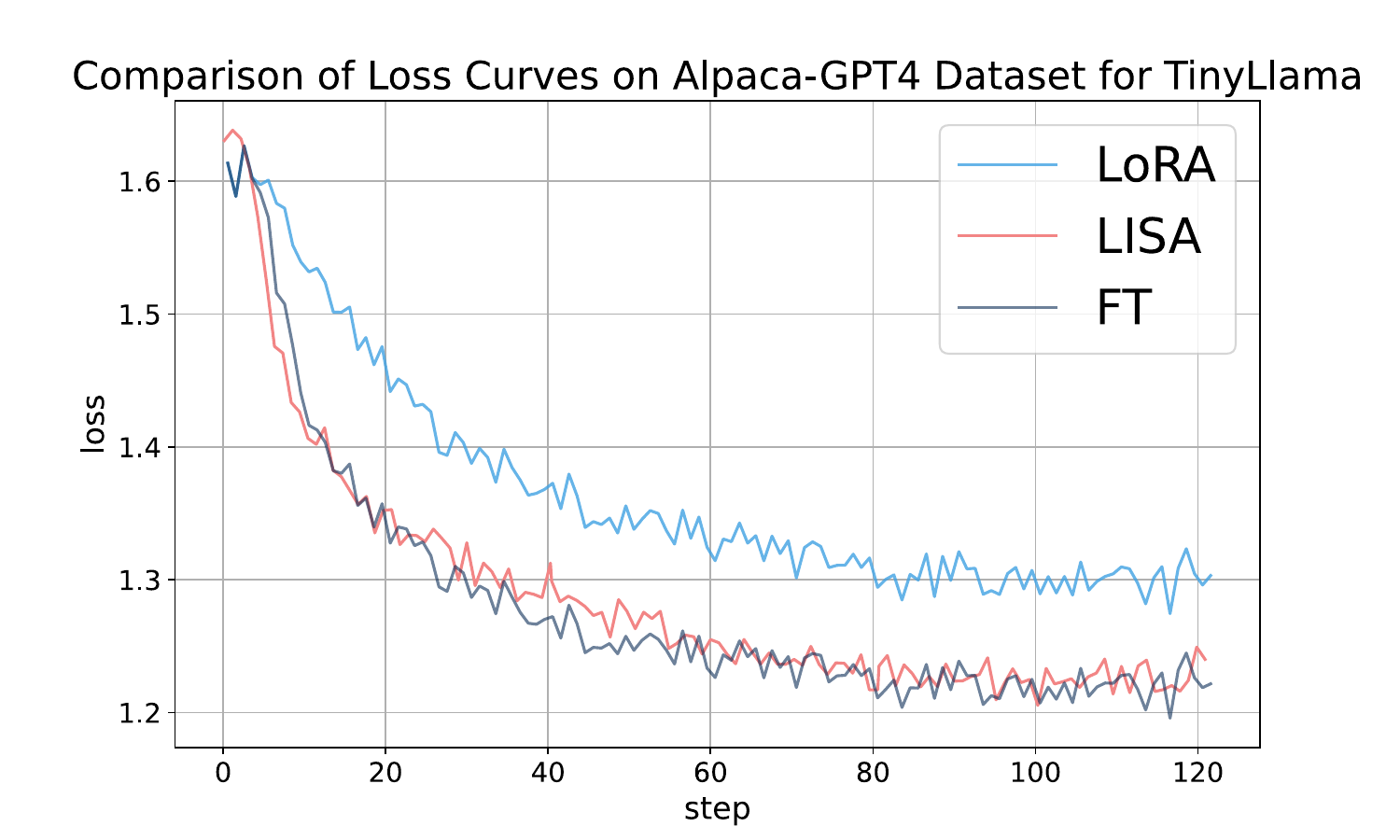}
\includegraphics[width=0.45\linewidth]{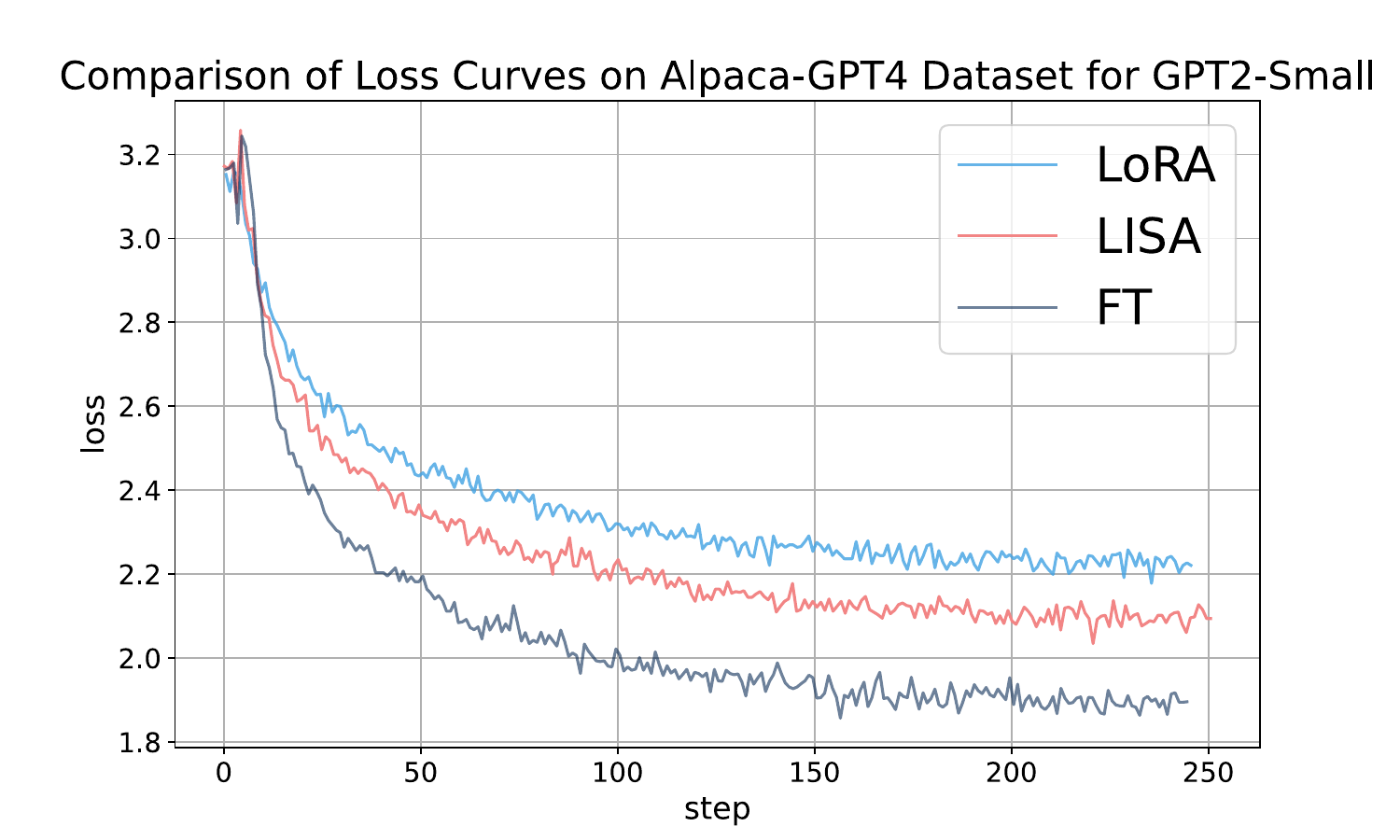}
\includegraphics[width=0.45\linewidth]{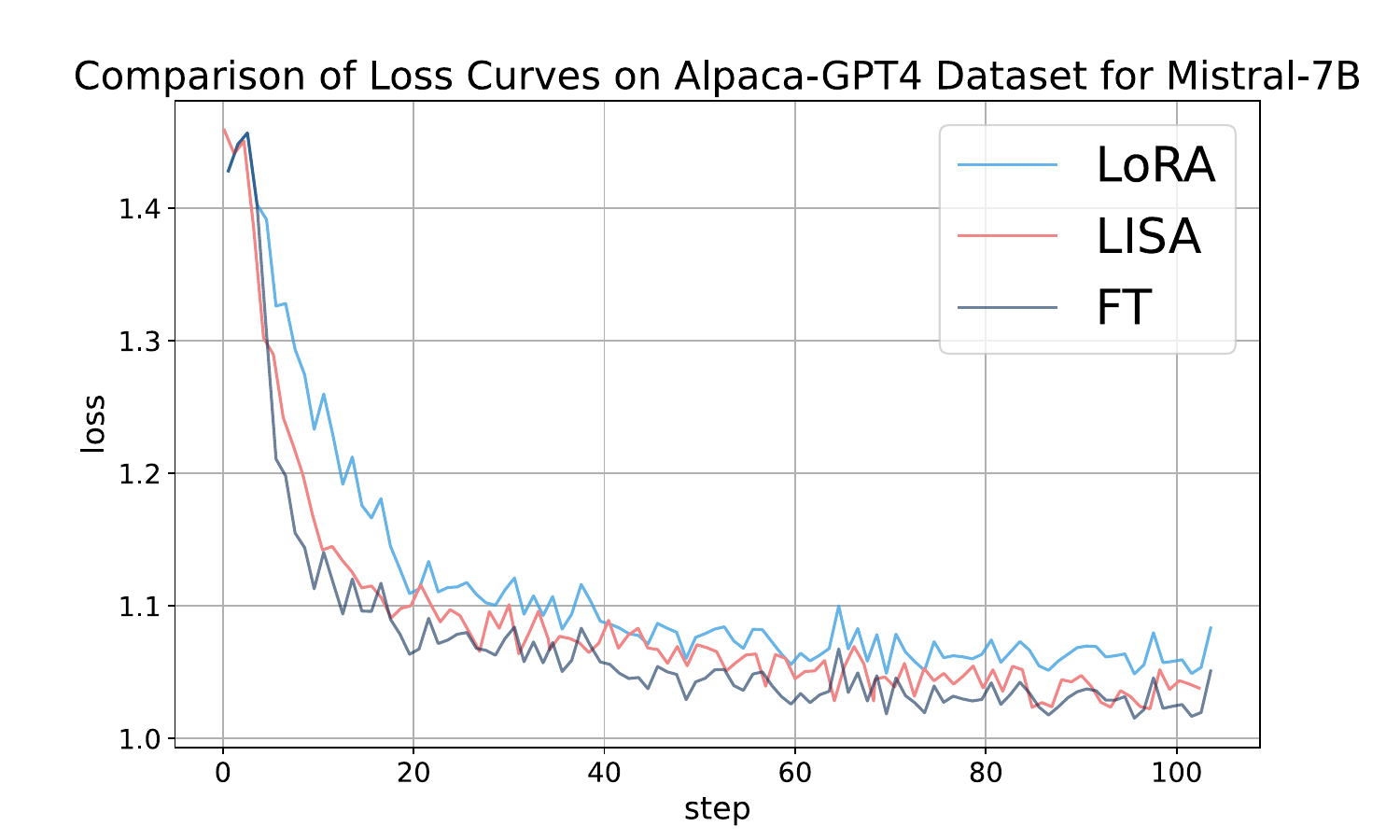}
\includegraphics[width=0.45\linewidth]{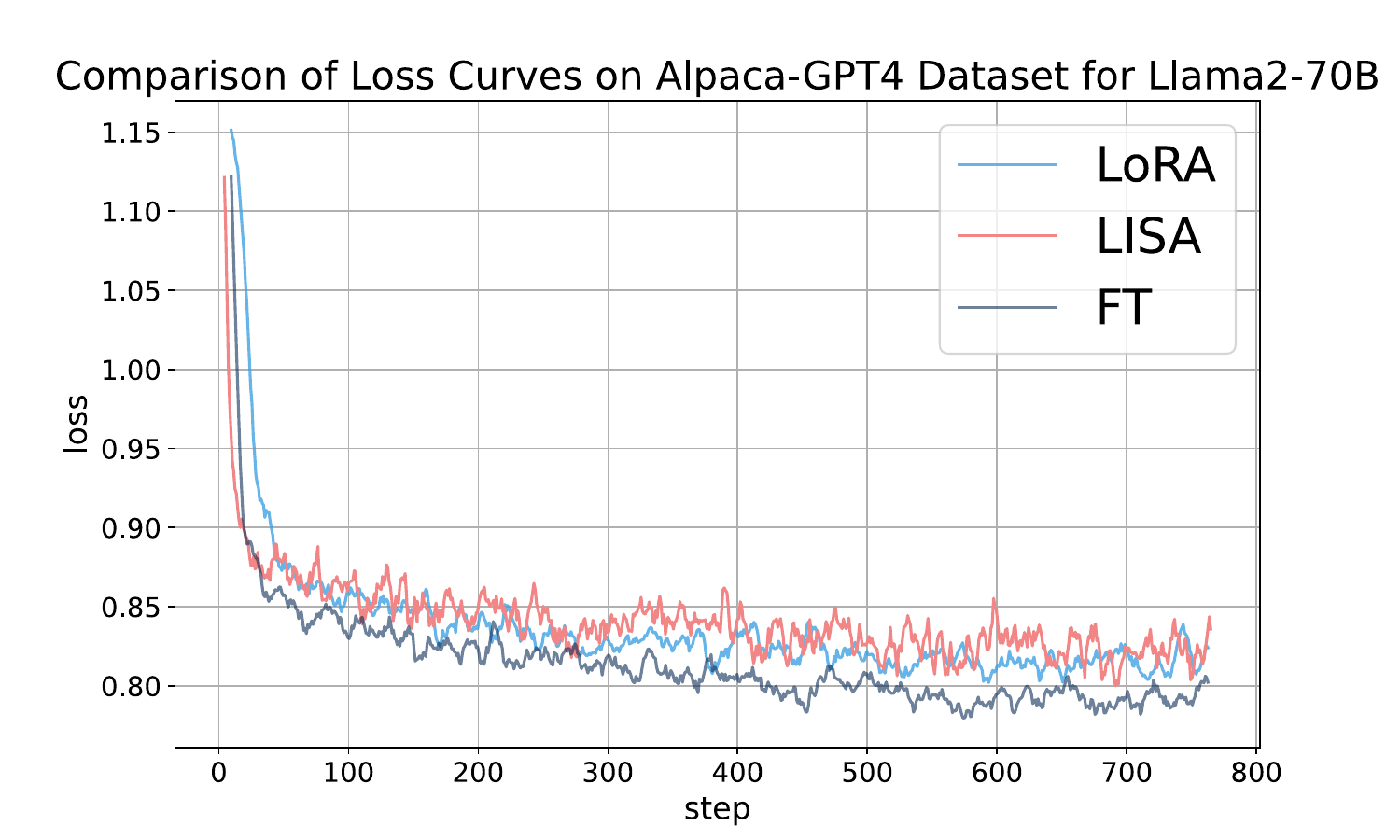}
\caption{Loss curves for LoRA, LISA, and full-parameter training on the Alpaca-GPT4 dataset across different models.}
\vspace{-0.2cm}

\label{fig:alpaca_gpt4_all_loss_curves}
\end{center}
\end{figure}

\begin{table*}[!ht]
\vspace{0.2cm}
\caption{Comparison of Language Model Fine-Tuning Methods on the MT-Bench score.}
\begin{center}
\begin{sc}
\centering
    \resizebox{\linewidth}{!}{
    \begin{tabular}{l|cccccccc|c}
        \toprule
                            & \multicolumn{9}{c}{MT-Bench}                                                                                                  \\
        Model \& Method     & Writing       & Roleplay      & Reasoning   & Code  & Math          & Extraction    & STEM          & Humanities    & Avg. $\uparrow$         \\ \midrule
        
        TinyLlama (Vanilla)
        & \cellcolor{lightblue}1.05
        & 2.25
        & 1.25
        & 1.00
        & 1.00
        & 1.00
        & \cellcolor{lightblue}1.45
        & \cellcolor{lightblue}1.00
        & 1.25
        \\


        TinyLlama (LoRA)
        & \cellcolor{lightblue}2.77
        & 4.05
        & 1.35
        & 1.00
        & \textbf{1.40}
        & 1.00
        & \cellcolor{lightblue}1.55
        & \cellcolor{lightblue}2.15
        & 1.90
        \\
        
        TinyLlama (GaLore)
        & \cellcolor{lightblue}\textbf{3.55}
        & \textbf{5.20}
        & 2.40
        & \textbf{1.15}
        & 1.40
        & \textbf{1.85}
        & \cellcolor{lightblue}2.95
        & \cellcolor{lightblue}2.40
        & \textbf{2.61}
        \\
        
        TinyLlama (\textbf{LISA})
        & \cellcolor{lightblue}3.30
        & 4.40
        & \textbf{2.65}
        & 1.12
        & 1.30
        & 1.75
        & \cellcolor{lightblue}\textbf{3.00}
        & \cellcolor{lightblue}\textbf{3.05}
        & \underline{$2.57$}
        \\ 
        
        \gr TinyLlama (FT)
        & \cellcolor{lightblue}3.27
        & 3.95
        & 1.35
        & 1.04
        & 1.33
        & 1.73
        & \cellcolor{lightblue}2.69
        & \cellcolor{lightblue}2.35
        & 2.21
        \\
        \midrule

         Mistral-7B (Vanilla)
         & \cellcolor{lightblue}5.25
         & 3.20
         & 4.50
         & 1.60
         & 2.70
         & \textbf{6.50}
         & \cellcolor{lightblue}\textbf{6.17}
         & \cellcolor{lightblue}4.65
         & 4.32
         \\

        
         Mistral-7B (LoRA)
         & \cellcolor{lightblue}5.30
         & 4.40
         & 4.65
         & \textbf{2.35}
         & \textbf{3.30}
         & 5.50
         & \cellcolor{lightblue}5.55
         & \cellcolor{lightblue}4.30
         & 4.41
         \\

         Mistral-7B (GaLore)
         & \cellcolor{lightblue}5.05
         & \textbf{5.27}
         & 4.45
         & 1.70
         & 2.50
         & 5.21
         & \cellcolor{lightblue}5.52
         & \cellcolor{lightblue}5.20
         & 4.36
         \\
         
         Mistral-7B (\textbf{LISA}) 
         & \cellcolor{lightblue}\textbf{6.84}
         & 3.65
         & \textbf{5.45}
         & 2.20
         & 2.75
         & 5.65
         & \cellcolor{lightblue}5.95
         & \cellcolor{lightblue}\textbf{6.35}
         & \underline{$\textbf{4.85}$}
         \\

         \gr Mistral-7B (FT)
         & \cellcolor{lightblue}5.50
         & 4.45
         & 5.45
         & 2.50
         & 3.25
         & 5.78
         & \cellcolor{lightblue}4.75
         & \cellcolor{lightblue}5.45
         & 4.64
         \\
         \midrule

         LLaMA-2-7B (Vanilla)
         & \cellcolor{lightblue}2.75
         & 4.40
         & 2.80
         & 1.55
         & 1.80
         & 3.20
         & \cellcolor{lightblue}5.25
         & \cellcolor{lightblue}4.60
         & 3.29
         \\

        
         LLaMA-2-7B (LoRA)
         & \cellcolor{lightblue}6.30
         & 5.65
         & \textbf{4.05}
         & 1.60
         & 1.45
         & 4.17
         & \cellcolor{lightblue}6.20
         & \cellcolor{lightblue}6.20
         & 4.45
         \\
         
         LLaMA-2-7B (GaLore)
         & \cellcolor{lightblue}5.60
         & 6.40
         & 3.20
         & 1.25
         & 1.95
         & \textbf{5.05}
         & \cellcolor{lightblue}6.57
         & \cellcolor{lightblue}7.00
         & 4.63
         \\
         LLaMA-2-7B (\textbf{LISA})
         & \cellcolor{lightblue}\textbf{6.55}
         & \textbf{6.90}
         & 3.45
         & \textbf{1.60}
         & \textbf{2.16}
         & 4.50
         & \cellcolor{lightblue}\textbf{6.75}
         & \cellcolor{lightblue}\textbf{7.65}
         & \underline{$\textbf{4.94}$}
         \\

         \gr LLaMA-2-7B (FT)
         & \cellcolor{lightblue}5.55
         & 6.45
         & 3.60
         & 1.75
         & 2.00
         & 4.70
         & \cellcolor{lightblue}6.45
         & \cellcolor{lightblue}7.50
         & 4.75
         \\
         
        \bottomrule
    \end{tabular}
    }
\end{sc}
\end{center}
\label{tab:all_mt_bench_full}
\end{table*}

Table~\ref{tab:all_mt_bench_full_70b} provides detailed MT-Bench scores for the LLaMA-2-70B model discussed in Section~\ref{sec:exp_large_ft}, demonstrating LISA's superior performance over LoRA in all aspects under large-scale training scenarios. Furthermore, in Figure~\ref{fig:alpaca_gpt4_all_loss_curves}, we observe that LISA consistently exhibits on-par or faster convergence speed than LoRA across different models, which provides strong evidence for LISA's superiority in practice.

\begin{table}[!htbp]
\caption{Mean score of three fine-tuning methods over three seeds for LLaMA-2-70B on the MT-Bench.}
\begin{center}
\begin{sc}
\centering
    \resizebox{\linewidth}{!}{
    \begin{tabular}{l|cccccccc|c}
        \toprule
                            & \multicolumn{9}{c}{MT-Bench}                                                                                                  \\
        Model \& Method     & Writing       & Roleplay      & Reasoning   & Code  & Math          & Extraction    & STEM          & Humanities    & Avg. $\uparrow$         \\ \midrule
        
        LLaMA-2-70B(Vanilla) 
        & 7.77
        & 5.52  
        & 2.95
        &1.70  
        & 1.70   
        & 6.40  
        & 7.42  
        & 8.07 
        & 5.19          \\

        LLaMA-2-70B(LoRA)    
        & 7.55
        & 7.00   
        & 5.30
        & 3.15
        & 2.60
        & 6.55   
        & 8.00   
        & 8.70  
        &   6.10        \\
        
        LLaMA-2-70B(\textbf{LISA})    
        & \textbf{8.18}
        & \textbf{7.90}
        & \textbf{5.45}
        & \textbf{4.45}
        & \textbf{2.75} 
        & \textbf{7.45}
        & \textbf{8.60} 
        & \textbf{9.05}
        & \textbf{6.72} \\ 

        \gr    LLaMA-2-70B(FT) 
        & 6.45 
        & 7.50   
        & 5.50
        & 3.40
        & 2.15  
        & 7.55
        & 8.10 
        & 9.40  
        & 6.25         \\
        
        \bottomrule
    \end{tabular}
    }
\end{sc}
\end{center}
\label{tab:all_mt_bench_full_70b}
\end{table}

The aforementioned results show that Vanilla LLaMA-2-70B excels in Writing, but full-parameter fine-tuning led to a decline in these areas, a phenomenon known as the ``Alignment Tax''~\citep{ouyang2022instructgpt}. This tax highlights the trade-offs between performance and human alignment in instruction tuning. LISA, however, maintains strong performance across various domains with a lower ''Alignment Tax``.


\newpage
\subsection{Continual Pre-training}
\begin{wrapfigure}{r}{0.45\textwidth}
\begin{center}
\vspace{-0.1 in}
\includegraphics[width=1.1\linewidth, trim=0 0 0 30]{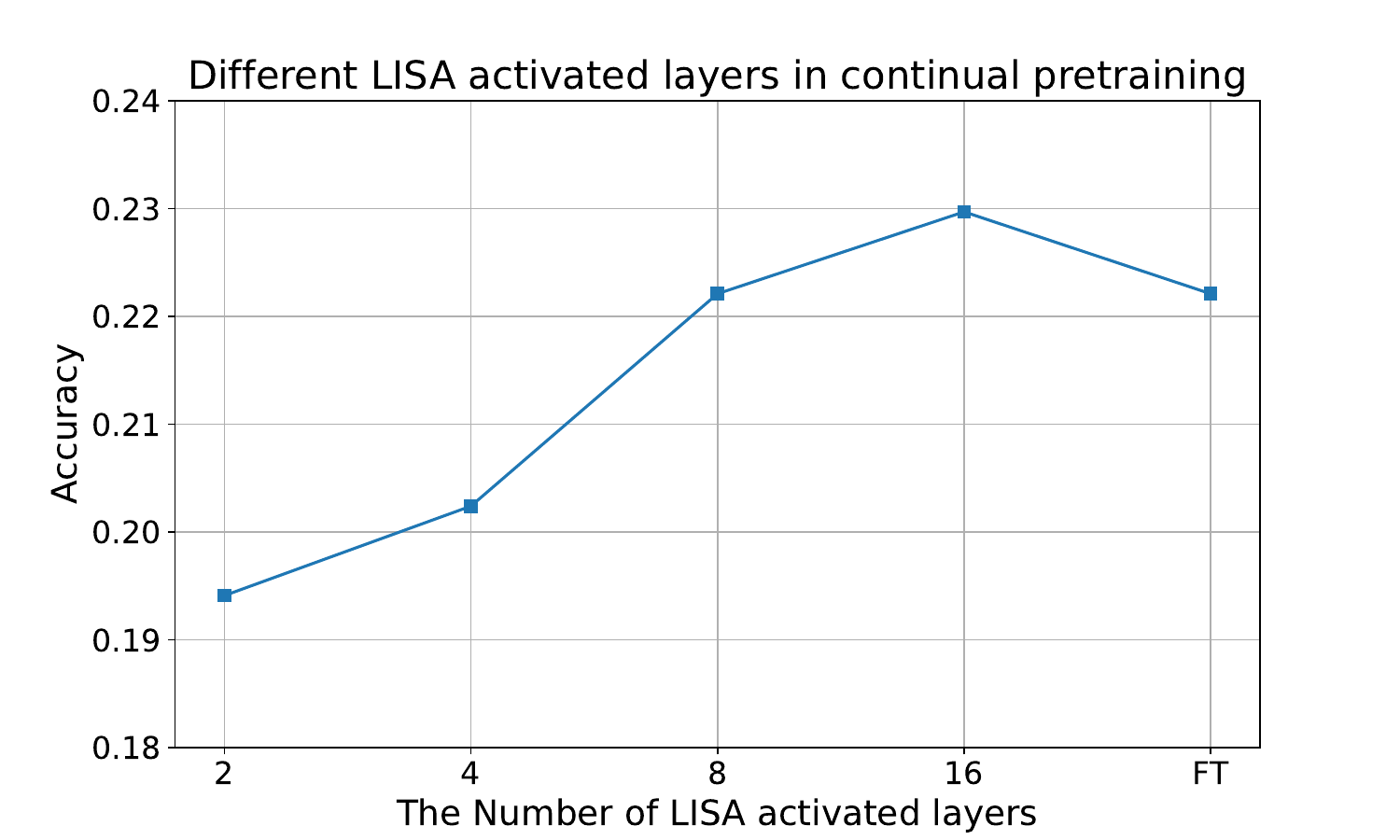}
\caption{The comparison of full parameter (FT) training and LISA with different sampling layers under continual pre-training scenario. The accuracy is the test set of GSM8K.}
\label{fig:con_pre_lisa_ft}
\end{center}
\vspace{-0.3 in}
\end{wrapfigure}

To better analyze the performance of LISA in the continual pre-training scenario, we used OpenWebMath for continual pre-training and the GSM8K train split for the fine-tuning stage, varying the number of sampling layers, $\gamma$, within LISA, ranging from ${2, 4, 8, 16}$, compared to the accuracy of the full parameter (FT) continual pre-training. Table \ref{fig:con_pre_lisa_ft} details the results for the LLaMA-2-7B model under various continual pre-training configurations. Notably, LISA with eight sampling layers achieves comparable accuracy with full parameters continual pre-training method. Furthermore, LISA with 16 sampling layers passes the accuracy of full parameter training. 

\subsection{Ablation Experiments}
\label{app:ablation}
\subsubsection{Sampling Layers $\gamma$}
\begin{wrapfigure}{r}{0.45\textwidth}
\begin{center}
\includegraphics[width=1.1\linewidth]{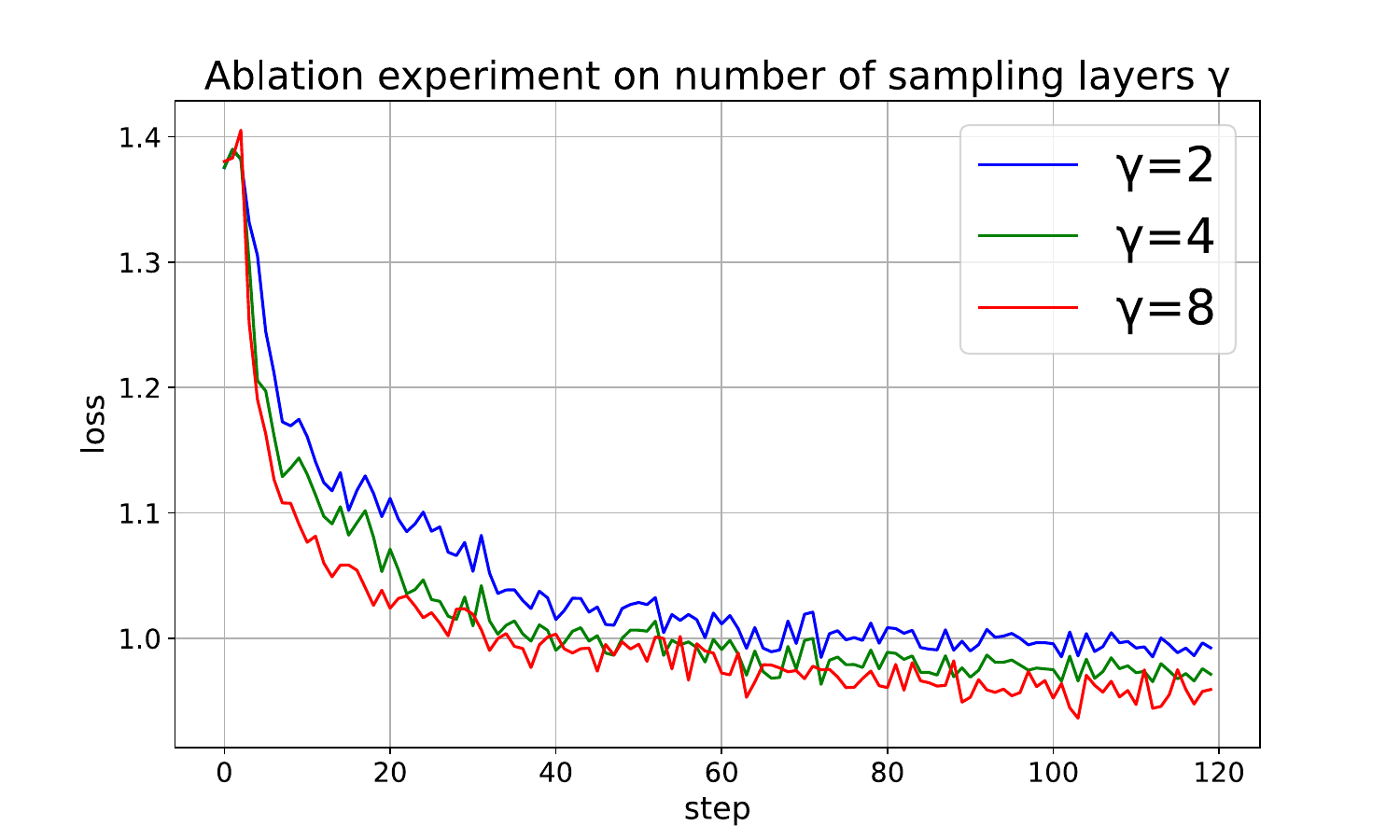}
\caption{Comparison of loss curves for the $\gamma$ ablation experiment.}
\label{fig:layers_ablation}
\end{center}
\end{wrapfigure}

We conducted an ablation study on the LLaMA-2-7B model trained with the Alpaca-GPT4 dataset, setting the sampling period $K=13$, so the number of samplings is exactly $10$. The study explored different configurations of sampling layers $\gamma$ including \{E+H+2L, E+H+4L, E+H+8L\}. Figure~\ref{fig:layers_ablation} depicts the impact of the number of sampling layers $\gamma$ on the training dynamics of the model. Three scenarios were analyzed: $\gamma=2$ (blue line), $\gamma=4$ (green line), and $\gamma=8$ (red line), throughout 120 training steps. Initially, all three configurations exhibit a steep decrease in loss, signaling rapid initial improvements in model performance. it's clear that the scenario with $\gamma=8$ consistently maintains a lower loss compared to the $\gamma=2$ and $\gamma=4$ configurations, suggesting that a higher $\gamma$ value leads to better performance in this context.

To better understand the LISA and hyperparameter Sampling Layers $\gamma$, we conducted ablation experiments on Sampling Layers $\gamma$ and learning rate $\eta$. The aim was to investigate the combined impact of these two variables on the LISA. Our experiments utilized the LLaMA-2-7B model, trained on the GSM8K dataset. We examined the effect of increasing the number of Sampling Layers $\gamma$ while simultaneously decreasing the learning rate $\eta$.

The Table \ref{tab:app_lr}, indicates that a higher number of sampling layers can enhance the model's effectiveness, provided the learning rate is adjusted appropriately. Specifically, the optimal performance is observed when the learning rate $\eta$ is reduced in proportion to the increase in the number of sampling layers $\gamma$, demonstrating the delicate balance required between these two parameters to maximize LISA's efficacy.
\begin{table}[!htbp]
\caption{Compare LISA with different Sampling Layers $\gamma$ and Learning Rate $\eta$, evaluate on GSM8K.}
\begin{sc}
\begin{center}
\centering
\resizebox{0.65\linewidth}{!}{
\begin{tabular}{c|cccc}
\toprule
 \multirow{2}[1]{*}{\makecell{Sampling \\ Layers $\gamma$}} & \multicolumn{4}{c}{Learning Rate  $\eta$} \\
& $5 \times 10^{-5}$ & $2.5 \times 10^{-5}$ & $1.25 \times 10^{-5}$ & $6.25 \times 10^{-6}$ \\ 
\midrule
2 & \textbf{15.77} & 15.32 & 15.21 & 15.01 \\
4 & 11.29 & 11.34 & \textbf{15.87} & 15.27 \\
8 & 13.32 & 14.39 & \textbf{16.30} & 15.32 \\
16 & 15.42 & 15.92 & 15.78 & \textbf{16.57} \\
\bottomrule
\end{tabular}
}
\end{center}
\label{tab:app_lr}
\end{sc}
\end{table}

\subsubsection{Sampling Period $K$}

\begin{wrapfigure}{r}{0.45\textwidth}
\vspace{-0.3 in}
\begin{center}
\includegraphics[width=1.1\linewidth]{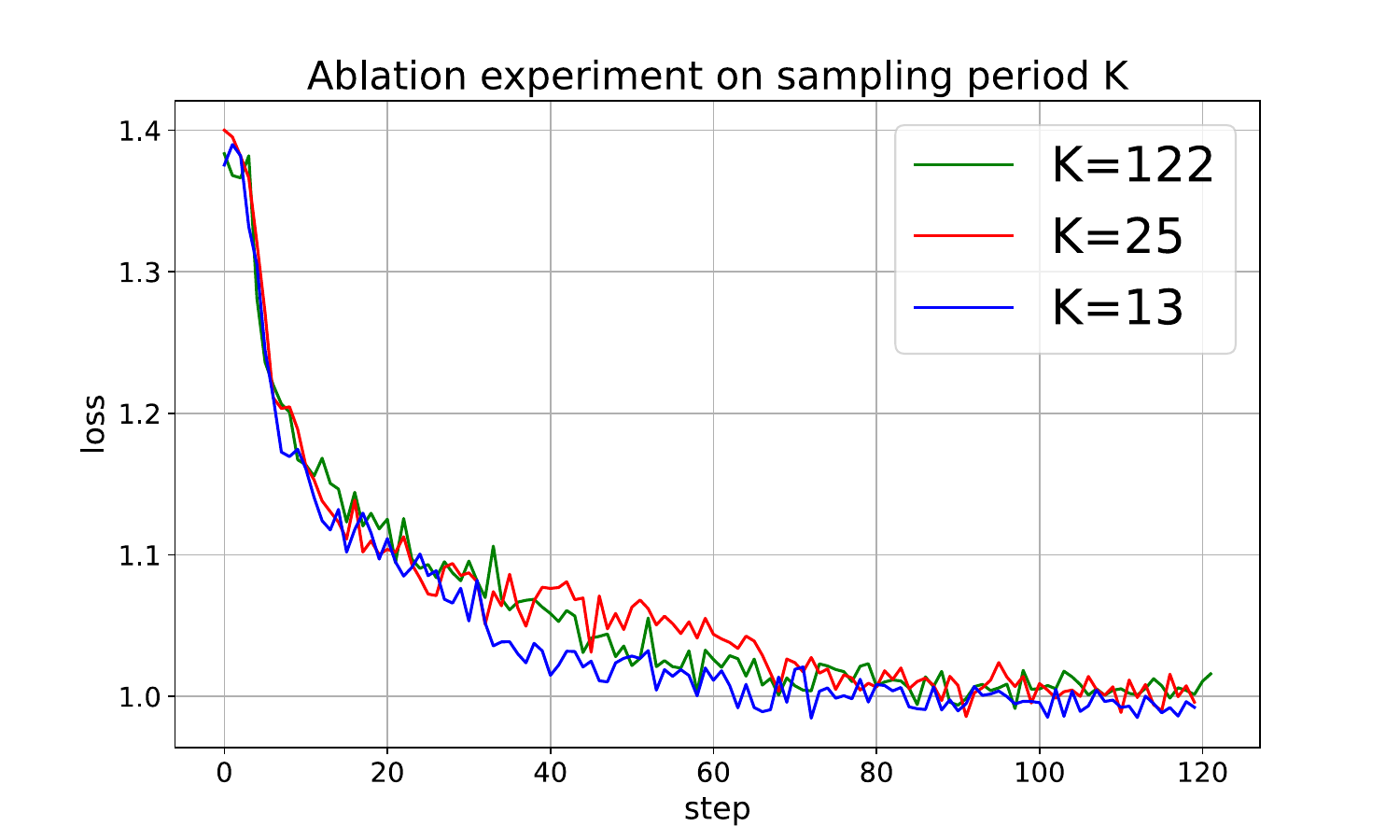}
\caption{Comparison of loss curves for the sampling period $K$ ablation experiment.}
\label{fig:K_ablation}
\vspace{-0.3 in}
\end{center}
\end{wrapfigure}

Figure~\ref{fig:K_ablation} displays the effects of varying sampling Period $K$ on training a 7B-sized model using the 52K-entry Alpaca-GPT4 dataset. This graph contrasts loss curves for different sampling period $K$ values: $K=122$ (green line), $K=25$ (red line), and $K=13$ (blue line) across 122 training steps. The results indicate that although each $K$ value results in distinct training trajectories, their convergence points are remarkably similar. This finding implies that for a 7B model trained on a 52K instruction conversation pair dataset, a sampling period of $K=13$ is optimal for achieving the best loss curve and corresponding MT-Bench score radar graph.

\subsubsection{Sensitiveness to Randomness}
\label{app:random}

\begin{wrapfigure}{r}{0.45\textwidth}
\vspace{-0.3 in}
\begin{center}
\includegraphics[width=1.1\linewidth]{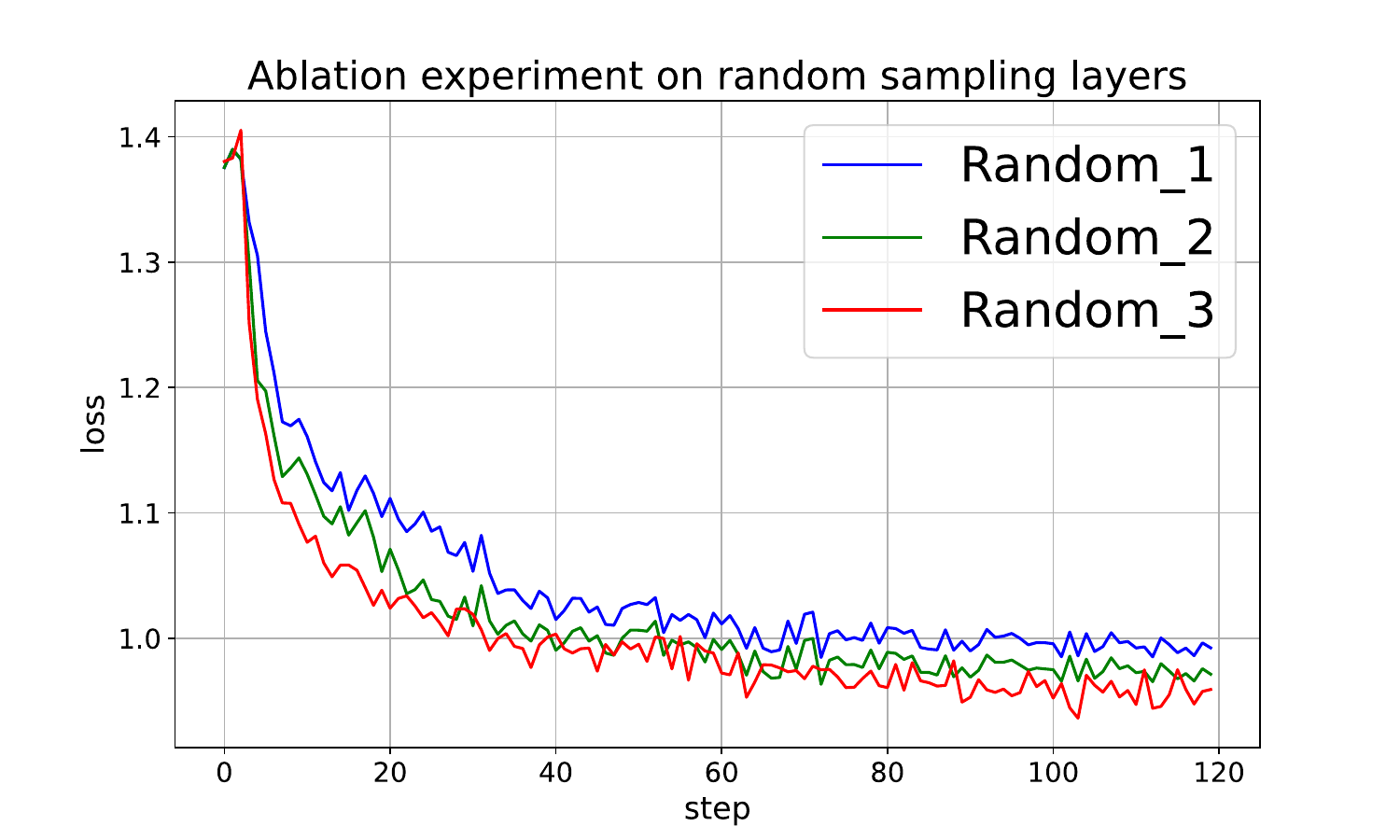}
\caption{Comparison of loss curves for random variance ablation experiment, indicating the loss metric over steps.}
\label{fig:random_variance}
\end{center}
\end{wrapfigure}

LLaMA-2-7B on Alpaca-GPT4 with update step per sampling period $K=13$, and sampling layers $\gamma=2$, run three times with different random layer pick. Figure~\ref{fig:random_variance} shows that different random selections of layers slightly affect the training process but converge similarly. Despite initial fluctuations, the loss trends of three runs—distinguished by blue, green, and red lines—demonstrate that the model consistently reaches a stable state, underscoring the robustness of the training against the randomness in layer selection.

Additionally, we also conduct experiments that analyze the impact of fixing randomly selected layers during training. The suggestion highlighted a need to evaluate the model's stability and performance under such conditions. To address this, we conducted further experiments using the LLaMA-2-7B model, maintaining identical hyperparameters and datasets as in our paper, and repeated the experiments three times to reduce variability from the random selection process.

The results are presented as table \ref{tab:app_fix_lisa}, with 'LISA-fix' denoting the experiments with randomly fixed layers and the appended number indicating the different selected seeds. The results are presented as table \ref{tab:app_fix_lisa}, with 'LISA-fix' denoting the experiments with randomly fixed layers and the appended number indicating the different selected seeds. 

\subsection{Performance with Early Exiting}

\begin{wraptable}{r}{0.3\textwidth}
\vspace{-0.3 in}
\caption{Compare LISA with fixed layers on LLaMA-2-7B, evaluate on MT-Bench.}
\begin{center}
\begin{sc}
\centering
\resizebox{\linewidth}{!}{
\begin{tabular}{l|c}
\toprule
 Method & MT-Bench $\uparrow$\\ \midrule

 LISA & \textbf{4.94} \\

   LISA-fix-1 & 4.62 \\

  LISA-fix-2 & 4.60 \\

  LISA-fix-3 & 4.67 \\

\bottomrule
\end{tabular}
}
\end{sc}
\end{center}
\label{tab:app_fix_lisa}
\end{wraptable}
According to previous works~\citep{chuang2023dola,fan2024not}, the early exiting strategy in LLMs is effective. We are interested in investigating whether the LISA algorithm will have a different impact when combined with the early exiting strategy. To explore this, we conducted a series of experiments using the LLaMA-2-7B model, focusing on various training methods and early exit points. The early exiting method is DoLa~\citep{chuang2023dola}.\newline

We conducted all of our experiments using the LLaMA-2-7B model, selecting layers \textbf{[0, 8, 16, 32]} as early exit points for evaluation under four different training conditions: Vanilla (baseline), Vanilla with DoLa, Full Parameter Fine-Tuning (FT) with DoLa, and LISA with DoLa. The model used is the same one trained on the Alpaca-GPT4 dataset, as reported in section \ref{sec:exp_moderate_ft}. This comprehensive approach enabled a detailed comparison of the model's performance and layer representation capabilities under various training methodologies. Due to time constraints, our experiments were conducted on a subset of 100 questions from the GSM8K test set, employing the same prompt as in the DoLa paper.

\begin{wraptable}{r}{0.3\textwidth}
\vspace{-0.3 in}
\caption{GSM8K Scores for LLaMA-2-7B when LISA meets the early exiting strategy DoLa.}
\begin{center}
\begin{sc}
\centering
\resizebox{\linewidth}{!}{
\begin{tabular}{l|c}
\toprule
 Method & GSM8K $\%$ $\uparrow$\\ \midrule

  Vanilla & 15 \\

  Vanilla + DoLa & 11 \\

  FT + DoLa & 16 \\

  \textbf{LISA} + DoLa & 17 \\

\bottomrule
\end{tabular}
}

\end{sc}
\end{center}
\label{tab:app_ee}
\end{wraptable}

From the table \ref{tab:app_ee}, it can be seen that the LISA algorithm does not negatively affect the representation or performance of some layers of the model; instead, it contributes to some improvements in effectiveness.

\subsection{Comparison of Evaluation Loss}
\begin{figure}[!htbp]
\vskip 0.2in
\begin{center}
\includegraphics[width=0.8\linewidth]{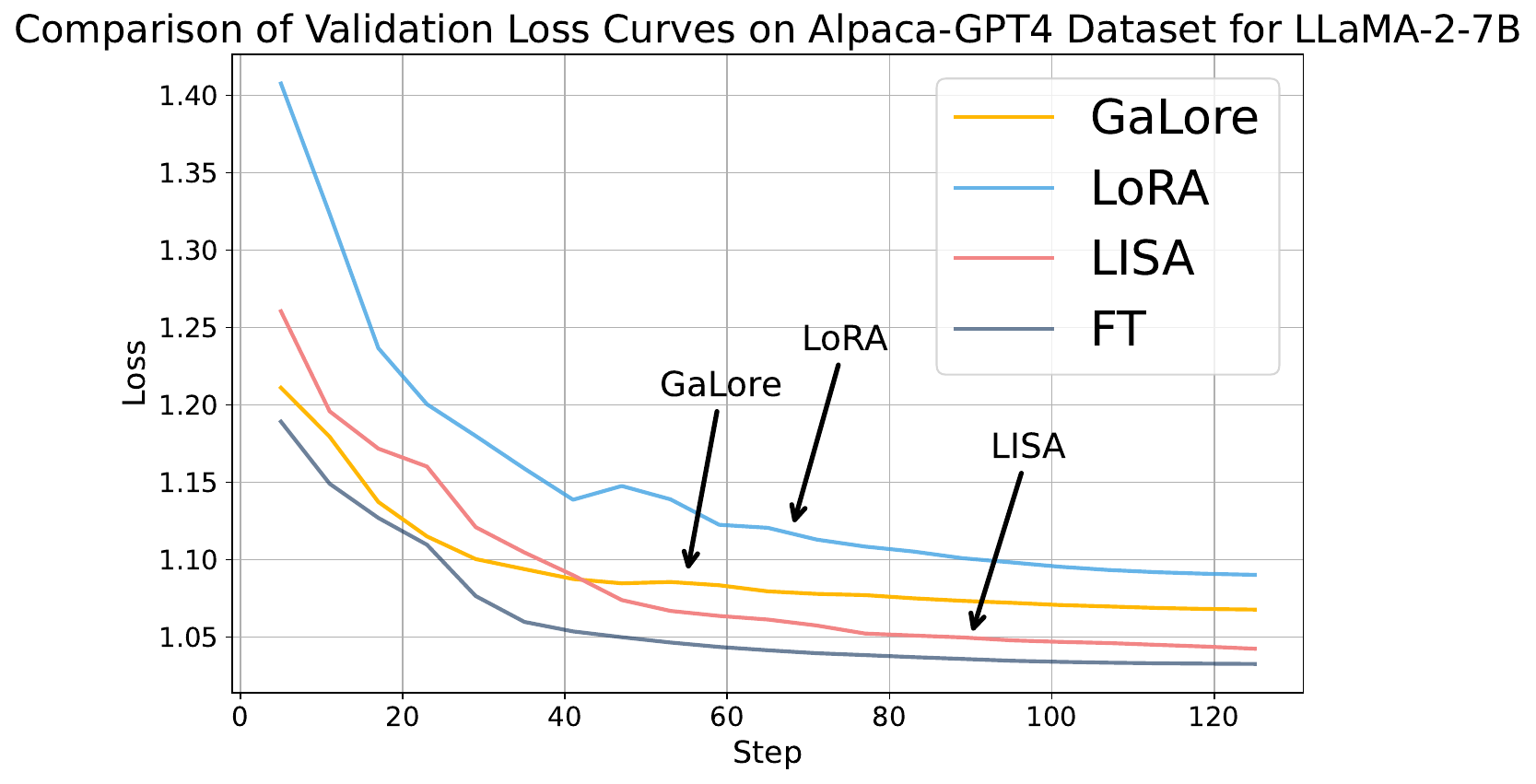}

\caption{Validation loss comparison on the Alpaca-GPT4 dataset for LLaMA-2-7B, showing LISA, GaLore, LoRA, and FT strategies, with arrows indicating specific observations in the loss trends.}
\label{fig:app_eval_loss}
\end{center}
\end{figure}

Besides the training loss, we also care about how LISA performs on the validation dataset. So, we split the Alpaca-GPT4 dataset into train and validation sets, the ratio is $9:1$, and ensure there is no data overlap between these two sets. Then we use the same setting in the sec \ref{sec:exp_moderate_ft}, training the LLaMA-2-7B on the Alpaca-GPT4 dataset with full parameter training (FT), LoRA, GaLore, and LISA. As Figure \ref{fig:app_eval_loss} shows, the trend in validation loss mirrors that observed in training loss, with LISA exhibiting no signs of overfitting. This consistency underlines LISA’s robustness in maintaining performance across different dataset splits.

\subsection{Additional Observations of Layerwise Skewness}
\label{appendix:weight_norm}

\begin{figure}[!htbp]
\vskip 0.2in
\begin{center}
\includegraphics[width=0.75\linewidth]{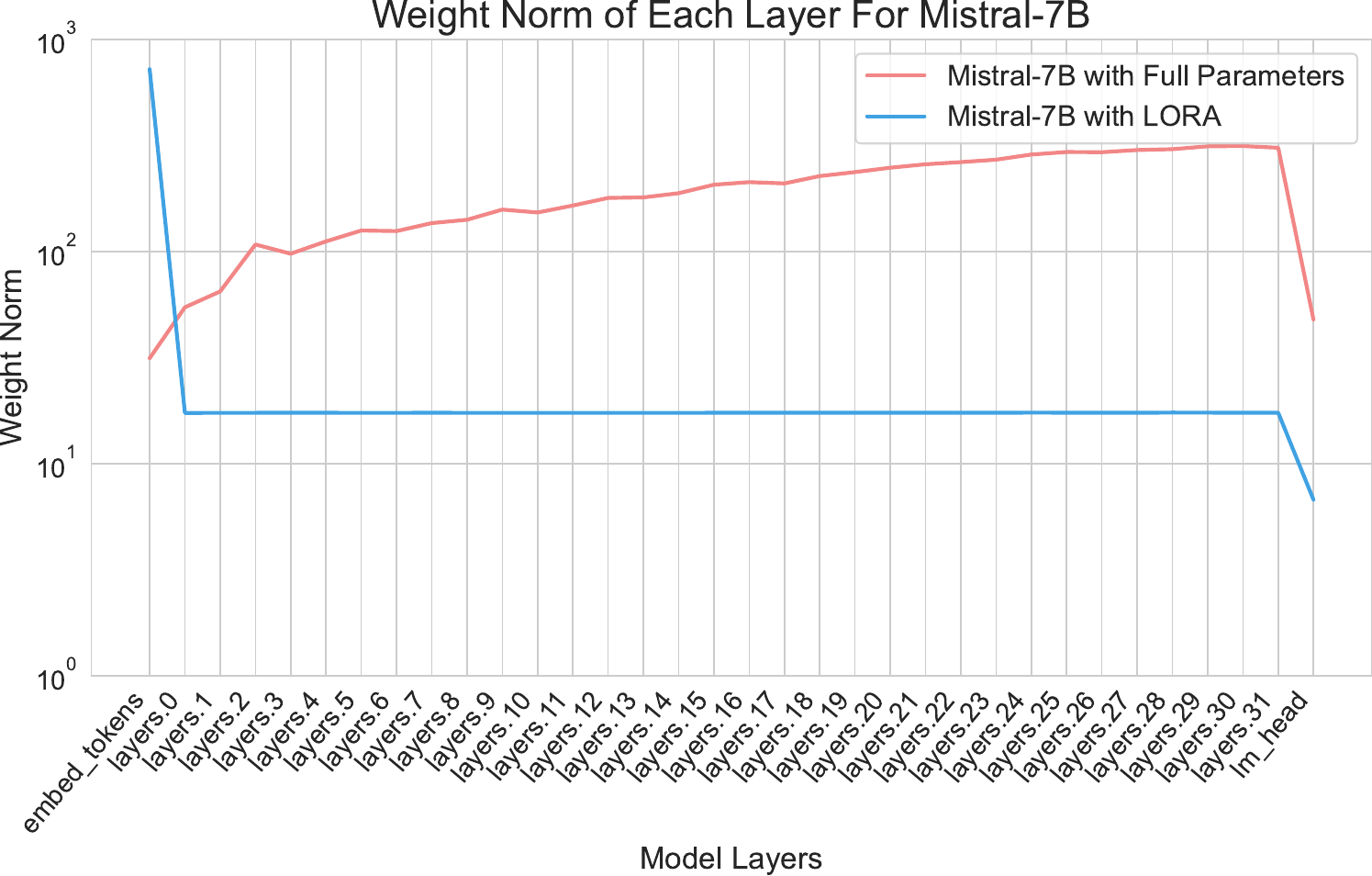}
\caption{Layer-wise weight norms during training of Mistral-7B with LoRA and Full Parameters training.}
\label{fig:app_weight_norm}
\end{center}
\end{figure}

We conduct further weight norm experiments on Mistral-7 B to support our motivation that the bottom and top layers have a more significant impact on the output.
Figure \ref{fig:app_weight_norm} provides similar observations as LLaMA-2-7B, where the bottom layer has a larger weight norm than other layers.

\section{Training Setup and Hyperparameters}
\label{appendix:train_setup_and_hyperparame}

\subsection{Training Setup}

In our experiments, we employ the LMFlow toolkit~\citep{diao2023lmflow}\footnote{https://github.com/OptimalScale/LMFlow} for conducting full parameter fine-tuning, LoRA tuning, and LISA tuning. We set the epoch number to 1 for fine-tuning and continual pre-training scenarios. Additionally, we utilized DeepSpeed offload technology~\citep{ren2021zerooffload} to run the LLMs efficiently. All experiments were conducted on $8\times$ NVIDIA Ampere Architecture GPU with 48 GB memory. Table \ref{tab:baseline_models} and table  \ref{tab:dataset_statistic} are the information covered in this paper.  

\begin{table}
\centering
\small
\caption{The statistics of datasets. \textsc{\# Train} and \textsc{\# Test} denote the number of training and test samples respectively. The unit for OpenWebMath is the number of documents.}
\vskip 0.15in
\begin{sc}
    \begin{tabular}{l|rr}
        \toprule
        Dataset  & \# Train & \# Test \\ \midrule
        
        Alpaca GPT-4\citep{peng2023instruction} 
        & 52,000        
        & -     \\

        MT-Bench~\citep{zheng2023judging} 
        & -        
        & 80     \\
        
        GSM8K~\citep{cobbe2021training}    
        & 7,473     
        & 1,319   \\
        
        MMLU~\citep{hendrycks2020measuring} 
        & - 
        & 14,079 \\

        AGIEval~\citep{zhong2023agieval}
        & -
        & 9316  \\

        WinoGrande~\citep{sakaguchi2021winogrande}
        & -
        & 44,000  \\
        
        PubMedQA~\citep{jin2019pubmedqa} 
        & 211,269   
        & 1,000   \\ 
        
        OpenWebMath~\citep{paster2023openwebmath} 
        & 6.3M   
        & -   \\ 
        
        \bottomrule
        \end{tabular}
\end{sc}
\label{tab:dataset_statistic}
\end{table}

\begin{wraptable}{r}{0.6\textwidth}
\centering
\small
\caption{Baseline Model Specifications}
\vskip 0.15in
\begin{sc}
    \resizebox{\linewidth}{!}{
    \begin{tabular}{l|rrrr}
        \toprule
        Model Name                      & \# Params & \# Layers & \makecell{Model\\Dim} & \# Heads \\
        \midrule
        \multicolumn{1}{l|}{TinyLlama}  & 1.1 B         & 22     & 2048      & 32    \\
        \multicolumn{1}{l|}{Mistral-7B} & 7 B           & 32     & 4096      & 32    \\
        \multicolumn{1}{l|}{LLaMA-2-7B}  & 7 B           & 32     & 4096      & 32    \\
        \multicolumn{1}{l|}{LLaMA-2-70B}  & 70 B           & 80     & 8192      & 64    \\
        \bottomrule
    \end{tabular}
}
\end{sc}
\label{tab:baseline_models}
\end{wraptable}

Our study explored a range of learning rates from $5 \times 10^{-6}$ to $3 \times 10^{-4}$, applying this spectrum to Full Parameter Training, LoRA, and LISA methods. For LoRA, we adjusted the rank $r$ to either 128 or 256 to vary the number of trainable parameters, applying LoRA across all linear layers. Regarding the number of sampling layers $\gamma$, our selections were guided by GPU memory considerations as reported in LoRA studies~\citep{hu2022lora}; For the LISA algorithm, we selected $\gamma=2$, and for experiments involving the 70B model, we opted for $\gamma=4$. The sampling period ($K$), defined as the number of update steps per sampling interval, ranges from 1 to 50. This range was influenced by variables such as the size of the dataset, the batch size, and the number of training steps. To manage this effectively, we partitioned the entire training dataset into $K$ segments, thereby enabling precise regulation of the training steps within each sampling period.

\subsection{Continual Pre-training Dataset}
\label{app:con_data}

We extracted a high-quality subset from OpenWebMath~\citep{paster2023openwebmath}, using the `\textit{Math\_score}' attribute from the metadata as the metric for high-quality instances. The `\textit{Math\_Score}' represents the probability that a document is mathematical, and we set the threshold at $0.95$. Finally, the number of tokens for this high-quality subset is 1.5 billion.

\subsection{Hyperparameter search}
\label{app:hpsearch}

We commenced our study with a grid search covering (i) learning rate, (ii) number of sampling layers $\gamma$, and (iii) sampling period $K$. Noting the effective performance of the LoRA method, we set the rank value to $r=128$ or $r=256$.

The optimal learning rate was explored within the range $\{5 \times 10^{-6}, 10^{-5}, 5 \times 10^{-5}, 6 \times 10^{-4}, 3 \times 10^{-4}\}$, applicable to full parameter training, LoRA, and LISA.  For GaLore, we adhered to the official Transformers implementation\footnote{https://huggingface.co/blog/galore}, utilizing default parameters, with the learning rate matching that of the full parameter training.

\begin{table}
\caption{The hyperparameter search identified optimal settings for each method: FP (Full Parameter Training), LoRA, GaLore, and LISA.}
\vskip 0.15in
\centering
    \begin{tabular}{ccccccc}
        \toprule
        \multicolumn{1}{c}{}               & \multicolumn{1}{c}{FP}   & \multicolumn{2}{c}{LoRA}                            & \multicolumn{3}{c}{LISA}                                                                        \\ \cmidrule(lr){2-2} \cmidrule(lr){3-4} \cmidrule(lr){5-7}
        \multicolumn{1}{c}{\textbf{Model}} & \multicolumn{1}{c}{lr}   & \multicolumn{1}{c}{lr}   & \multicolumn{1}{c}{Rank} & \multicolumn{1}{c}{lr}   & \multicolumn{1}{c}{$\gamma$} & \multicolumn{1}{c}{$K$}    \\ \cmidrule(lr){1-1} \cmidrule(lr){2-2} \cmidrule(lr){3-4} \cmidrule(lr){5-7}
        \multicolumn{1}{c}{GPT2-Small}     & \multicolumn{1}{c}{$3 \times 10^{-4}$} & \multicolumn{1}{c}{$6 \times 10^{-4}$} & \multicolumn{1}{c}{128}  & \multicolumn{1}{c}{$6 \times 10^{-4}$} & \multicolumn{1}{c}{2}                     & \multicolumn{1}{c}{3} \\
        TinyLlama                          & $5 \times 10^{-6}$                    & $5 \times 10^{-5}$                     & 128                      & $5 \times 10^{-5}$                     & 2                                         & 10                     \\
        Mistral-7B                         & $5 \times 10^{-6}$                     & $5 \times 10^{-5}$                     & 128                      & $5 \times 10^{-5}$                     & 2                                         & 10                     \\
        LLaMA-2-7B                          & $5 \times 10^{-6}$                     & $5 \times 10^{-5}$                     & 128                      & $5 \times 10^{-5}$                     & 2                                         & 10                    \\
        LLaMA-2-70B                         & $5 \times 10^{-6}$                     & $5 \times 10^{-5}$                     & 128                      & $5 \times 10^{-5}$                     & 4                                         & 10         \\
        \bottomrule
    \end{tabular}
\label{tab:hp-search-results}
\end{table}

Regarding the number of sampling layers $\gamma$, in alignment with Table~\ref{tab:peak_gpu_memory}, we selected values that matched or were lower than LoRA's GPU memory cost. Consequently, $\gamma=2$ was predominantly used in the LISA experiments, while $\gamma=4$ was chosen for the 70B model experiments.

For the sampling period $K$, we examined values within ${1,3,5,10,50,80}$, aiming to maintain the model's update steps within a range of 10 to 50 per sampling period. This selection was informed by dataset size, batch size, and total training steps.

The comprehensive results of our hyperparameter search, detailing the optimal values for each configuration, are presented in Table \ref{tab:hp-search-results}.

\section{Licenses}
\label{app_licenses}
For instruction following and domain-specific fine-tuning tasks, all the datasets, including Alpaca~\citep{alpaca}, GSM8k~\citep{cobbe2021training}, MMLU~\citep{hendrycks2020measuring}, AGIEval~\citep{zhong2023agieval} and PubMedQA~\citep{jin2019pubmedqa} are released under MIT license. WinoGrande~\citep{sakaguchi2021winogrande} and MT-Bench~\citep{zheng2023judging} are under Apache-2.0 license.  For GPT-4, the generated dataset is only for research purposes, which shall not violate its terms of use.

\end{document}